\documentclass[onefignum,onetabnum]{siamart171218}

\usepackage{lipsum}
\usepackage{amsfonts}
\usepackage{graphicx}
\usepackage{epstopdf}
\usepackage{algorithmic}
\ifpdf
  \DeclareGraphicsExtensions{.eps,.pdf,.png,.jpg}
\else
  \DeclareGraphicsExtensions{.eps}
\fi

\newsiamremark{remark}{Remark}
\newsiamremark{hypothesis}{Hypothesis}
\crefname{hypothesis}{Hypothesis}{Hypotheses}
\newsiamthm{claim}{Claim}

\headers{NeuralUQ}{Z. Zou, X. Meng, A. F. Psaros, G. E. Karniadakis}

\title{NeuralUQ: A comprehensive library for uncertainty quantification in neural differential equations and operators}

\author{Zongren Zou\thanks{Division of Applied Mathematics, Brown University, Providence, RI 02912, USA.}
\and Xuhui Meng\thanks{Institute of Interdisciplinary Research for Mathematics and Applied Science, School of Mathematics and Statistics, Huazhong University of Science and Technology, Wuhan 430074, China. (Corresponding author: \email{xuhui\_meng@hust.edu.cn}).}
\and Apostolos F Psaros\footnotemark[1]
\and George Em Karniadakis$^*$\thanks{Pacific Northwest National Laboratory, Richland, WA 99354, USA.}}

\usepackage{amsopn}

\DeclareMathAlphabet{\pazocal}{OMS}{zplm}{m}{n}

\newcommand{\cD}{\pazocal{D}}

\newcommand{\cM}{\pazocal{M}}

\usepackage{multirow} 

\usepackage{listings}
\usepackage{color}

\definecolor{dkgreen}{rgb}{0,0.6,0}
\definecolor{gray}{rgb}{0.5,0.5,0.5}
\definecolor{mauve}{rgb}{0.58,0,0.82}

\usepackage{bm}
\usepackage{subfigure}
\usepackage{bold-extra}
\usepackage{booktabs}
\usepackage{mathtools}
\floatname{algorithm}{Procedure}
\graphicspath{{./figs/}}

\ifpdf
\hypersetup{
  pdftitle={NeuralUQ},
  pdfauthor={Z. Zou, X. Meng, A. F Psaros, G. E. Karniadakis}
}
\fi

\begin{document}

\maketitle



\begin{abstract}
Uncertainty quantification (UQ) in machine learning is currently drawing increasing research interest, driven by the rapid deployment of deep neural networks across different fields, such as computer vision, natural language processing, and the need for reliable tools in risk-sensitive applications. 
Recently, various machine learning models have also been developed to tackle problems in the field of scientific computing with applications to computational science and engineering (CSE). 
Physics-informed neural networks and deep operator networks are two such models for solving partial differential equations and learning operator mappings, respectively.
In this regard, a comprehensive study of UQ methods tailored specifically for scientific machine learning (SciML) models has been provided in \cite{psaros2022uncertainty}. 
Nevertheless, and despite their theoretical merit, implementations of these methods are not straightforward, especially in large-scale CSE applications, hindering their broad adoption in both research and industry settings.
In this paper, we present an open-source Python library 
(\href{https://github.com/Crunch-UQ4MI}{\textit{github.com/Crunch-UQ4MI}}),
termed {\emph{NeuralUQ}} and accompanied by an educational tutorial, for employing UQ methods for SciML in a convenient and structured manner. 
The library, designed for both educational and research purposes, supports multiple modern UQ methods and SciML models. It is based on a succinct workflow and facilitates flexible employment and easy extensions by the users.
We first present a tutorial of NeuralUQ and subsequently demonstrate its applicability and efficiency in four diverse examples, involving dynamical systems and high-dimensional parametric and time-dependent PDEs. 
\end{abstract}



\section{Introduction}\label{sec:intro}

Physics-informed and more generally scientific machine learning (SciML) provide potent tools for accurately modeling and predicting the dynamic response of physical systems; see \cite{karniadakis2021physicsinformed,cuomo2022scientific, willard2021integrating} for a comprehensive review as well as \cite{cai2022physics,psaros2022uncertainty}.
Neural networks (NNs) for solving ordinary/partial differential equations (ODEs, PDEs), such as the physics-informed NN (PINN), and for learning operator mappings, such as the deep operator network (DeepONet), seamlessly combine scattered observational  data with available physical models (e.g. \cite{lye2020deep, kashinath2021physics, trask2022enforcing, rao2020physics, sun2020surrogate}).
Specifically, they can address mixed
problems when only partial observational data and information regarding the underlying physical system
are available (e.g., \cite{raissi2020hidden,cai2021flow}); it is straightforward to incorporate noisy and multi-fidelity data (e.g., \cite{yang2021bpinns,meng2022learning,meng2020composite,meng2021multifidelity});
they can provide physically consistent predictions even for extrapolatory tasks (e.g., \cite{yang2021bpinns}); they do not
require computationally expensive mesh generation (e.g., \cite{raissi2019physicsinformed,lu2021deepxde}); and open-source software is available and
continuously extended (e.g., \cite{lu2021deepxde,hennigh2021nvidia, koryagin2019pydens,chen2020neurodiffeq, rackauckas2019diffeqflux, rackauckas2020universal, haghighat2021sciann, xu2020adcme}). 

\begin{figure}[!ht]
	\centering
	\includegraphics[width=0.95\linewidth]{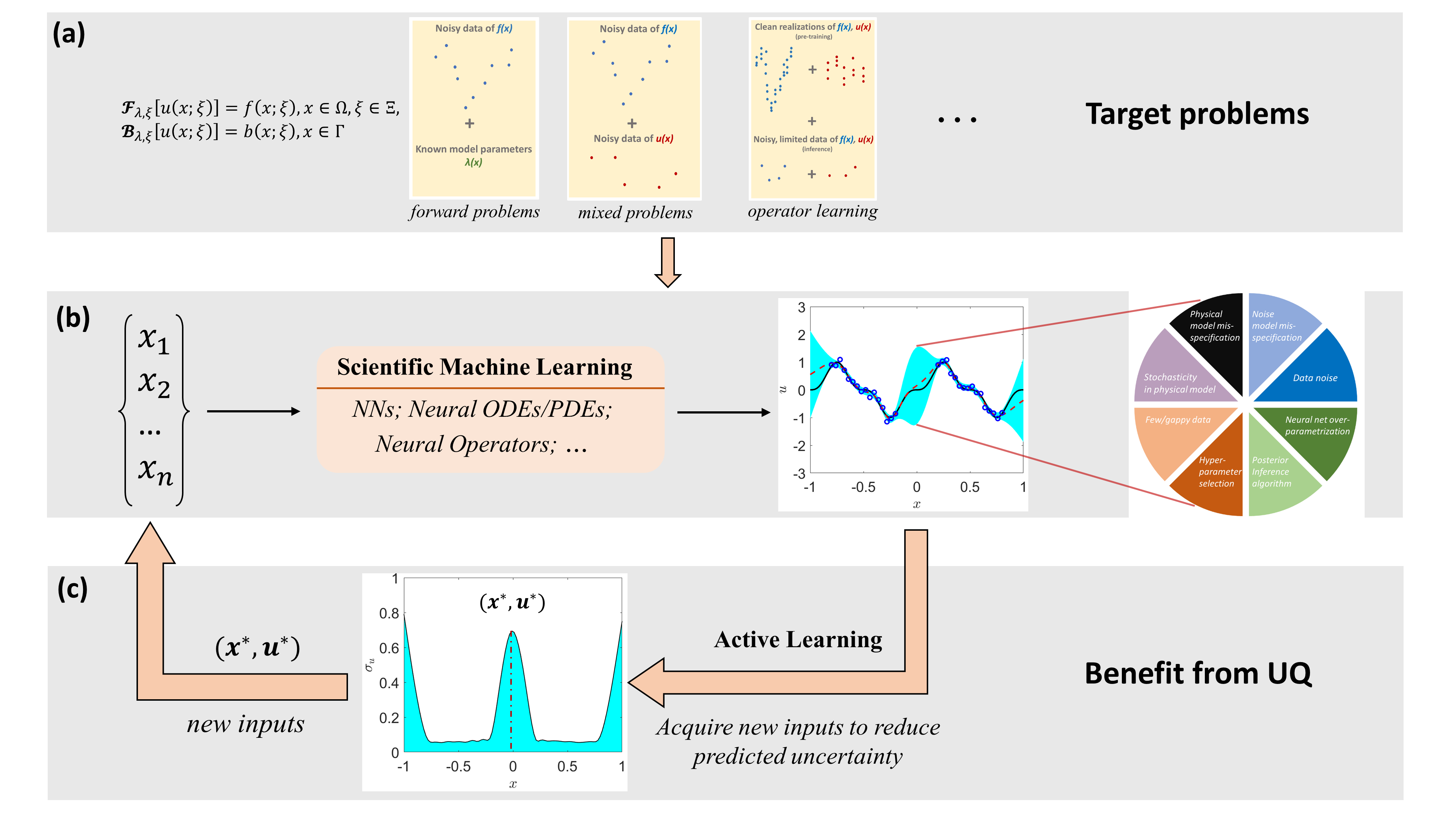}
	\caption{\label{fig:sciml_pie}
	Schematic overview of a SciML problem scenario.
	As shown in (a), SciML problems of interest include, but are not limited to, forward, inverse and mixed ODEs/PDEs, and operator learning problems. 
	Next, in the pie chart of (b) a qualitative breakdown of total uncertainty is shown, describing the contributions from data (noisy, gappy); physical models (misspecification, stochasticity); neural networks (architecture, hyperparameters, overparametrization); and posterior inference. 
	Note that we mainly address two types of uncertainty in the current paper, i.e., aleatoric (data noise) and  epistemic (NN overparametrization and limited data) uncertainties. 
	As shown in (c), active learning for improving the predictions can subsequently be performed, i.e., to acquire new inputs according to an acquisition function that is based on the output uncertainty from part (b).  
}
\end{figure}

However, uncertainty, associated with noisy and limited data as well as NN
overparameterization, for example, can degrade the accuracy of these models significantly \cite{abdar2021review,psaros2022uncertainty, tran2016edward, gardner2018gpytorch}.
As a result, uncertainty quantification (UQ) is paramount for deep learning to be \textit{reliably} used in critical applications involving
physical and biological systems \cite{wilson2020case,qin2021deep,tipireddy2021time,babaee2020multifidelity, zhu2019physics, tripathy2018deep}. 
In Fig.~\ref{fig:sciml_pie}, we provide an overview of SciML problems of interest as well as different sources of uncertainty.
In addition to reliability concerns, UQ is also necessary for efficient and economical design; for capital budgeting and risk-adaptive decision making in related projects; and even for assessing investment opportunities involving such systems in the form of real options \cite{richard2006real,de2008real}.
Clearly, this broad view of applications gives to UQ an additional dimension pertaining to the potential of \textit{harnessing uncertainty}, and renders it an indispensable research area of SciML applied to computational science and engineering problems and beyond.


Further, UQ in SciML is considerably more challenging than in traditional numerical solvers because of optimization-related issues of over-parameterized NNs, as well as the various uncertainties shown in Fig.~\ref{fig:sciml_pie}.
In addition, SciML is applied to increasingly more diverse problems, from steady to time-dependent and high-dimensional ones, and to problems involving systems given only in the form of black-box models (see, e.g., \cite{yang2021bpinns,meng2022learning,psaros2022uncertainty,han2018solving,darbon2020overcoming}).
Moreover, evaluating the performance of the available UQ methods is not straightforward, especially in situations involving limited data.
In this regard, Psaros et al. \cite{psaros2022uncertainty} provided an extensive review and proposed novel methods of UQ for SciML.
Nevertheless, employing current methods in large-scale applications and developing new ones, where trustworthy implementations and fast experimentation are needed, requires an up-to-date, well-maintained, and versatile software package.
To this end, we present in this paper an open-source Python library termed NeuralUQ, which can be accessed at 
\href{https://github.com/Crunch-UQ4MI}{\textit{github.com/Crunch-UQ4MI}}.
We expect that NeuralUQ will be a useful toolbox for both practitioners and researchers, who either choose to apply off-the-shelf methods for solving their problems, or to build their own methods based on the ones the library currently supports.
In passing, we note that NeuralUQ is constantly updated and its most recent capabilities can be viewed in the GitHub repository.

We organize the paper as follows. 
In Section~\ref{sec:uqINsciml}, we briefly present SciML, specifically the PINN and DeepONet methods, as well as recent advances in UQ for SciML, which are the building blocks for the library; however, other neural models for PDEs and operators can be employed as well. 
In Section~\ref{sec:library}, we describe the design of the NeuralUQ library, which includes model construction, posterior inference, and performance evaluation as well as post-training calibration.
In Section~\ref{sec:examples}, we provide four computational examples for demonstrating the applicability and capabilities of our library.
Lastly, in Section~\ref{sec:conclusions}, we conclude with the findings of this work and discuss future research.

\section{Uncertainty quantification for scientific machine learning}\label{sec:uqINsciml}

In this section, we first briefly describe various SciML problem scenarios and the recently developed deep learning methods for solving them.  Subsequently, we review the related UQ methods as well as the approaches for model performance evaluation and for post-training calibration.  We also provide 
a glossary table of frequently used terms in this paper, and in ML/SciML more generally, in Appendix~\ref{sec:glossary}. 
The interested reader is directed to \cite{psaros2022uncertainty} for more detailed information regarding UQ methods for SciML.

\subsection{Scientific machine learning}
\label{sec:uqINsciml:sciml}

Consider a nonlinear ODE/PDE describing the dynamics of a physical system as follows:
\begin{subequations}\label{eq:intro:piml:pinn:pde}
	\begin{align}
		\pazocal{F}_{\lambda, \xi}[u(x; \xi)] &= f(x; \xi) \text{, } x \in  \Omega\text{, } \xi \in \Xi, 
		\label{eq:intro:piml:pinn:pde:a}\\
		\pazocal{B}_{\lambda, \xi}[u(x; \xi)] & = b(x; \xi) \text{, } x \in \Gamma, 
		\label{eq:intro:piml:pinn:pde:b}
	\end{align}
\end{subequations}
where $x$ is the $D_x$-dimensional space-time coordinate, $\Omega$ is a bounded domain with boundary $\Gamma$, $\xi$ is a random event in a probability space $\Xi$, while $f(x; \xi)$ and $u(x; \xi)$ represent the $D_u$-dimensional source term and sought solution evaluated at $(x, \xi)$, respectively. 
Further, $\pazocal{F}_{\lambda, \xi}$ is a general differential operator; $\pazocal{B}_{\lambda, \xi}$ is a boundary/initial condition operator acting on the domain boundary $\Gamma$; and $b$ is a boundary/initial condition term.
Furthermore, $\lambda(x; \xi)$, with $x \in \Omega_{\lambda} \subset \Omega$, denotes the problem parameters.
We refer to problems where the operators $\pazocal{F}_{\lambda, \xi}$ and $\pazocal{B}_{\lambda, \xi}$ are known, i.e., the differential operators that define them are specified, as \textit{neural ODEs/PDEs}, because we use neural networks (NNs) throughout the present work to solve them.
We refer to problems involving unknown operators $\pazocal{F}_{\lambda, \xi}$ and $\pazocal{B}_{\lambda, \xi}$ as \textit{neural operators}, because we use NNs to learn them.

For neural ODEs/PDEs, $\pazocal{F}_{\lambda, \xi}$, $\pazocal{B}_{\lambda, \xi}$ are known and deterministic, i.e. $\xi$ is fixed, while $\lambda(x)$, and $u(x)$ are partially unknown. Given noisy data of $u$ and $\lambda$ sampled at random finite locations in $\Omega$ and $\Omega_{\lambda}$, respectively, as well as noisy data of $f$ and $b$, the objective of this problem is to obtain the solution $u(x)$ at every $x \in \Omega$ and the problem parameters $\lambda(x)$ at every $x \in \Omega_{\lambda}$.
The dataset in this case is expressed as $\cD = \{\cD_f, \cD_b, \cD_u, \cD_{\lambda}\}$, where $\cD_u = \{x_i, u_i\}_{i=1}^{N_u}$, $\cD_{\lambda} = \{x_i, \lambda_i\}_{i=1}^{N_{\lambda}}$, $u_i = u(x_i)$, $\lambda_i = \lambda(x_i)$, and $\cD_f, \cD_b$ are given as above.
The PINN method, developed in \cite{raissi2019physicsinformed}, addresses the PDE problems deterministically, by constructing NN approximators, parameterized by $\theta$, for modeling the solution $u(x)$.
The approximator $u_{\theta}(x)$ is substituted into Eq.~\eqref{eq:intro:piml:pinn:pde} via automatic differentiation \cite{tensorflow2015-whitepaper} for producing $f_{\theta} = \pazocal{F}_{\lambda, \xi}[u_{\theta}]$ and $b_{\theta} = \pazocal{B}_{\lambda, \xi}[u_{\theta}]$.
For solving the mixed deterministic PDE problem, an additional NN can be constructed for modeling $\lambda(x)$.
Finally, the optimization problem for obtaining $\theta$ in the mixed problem scenario with the dataset $\cD = \{\cD_f, \cD_b, \cD_u, \cD_{\lambda}\}$ is cast as
\begin{multline}\label{eq:intro:piml:pinn:loss:1}
	\hat{\theta} = \underset{\theta}{\mathrm{argmin}}~\pazocal{L}(\theta) \text{, where } 
	\pazocal{L}(\theta) = \frac{w_f}{N_f}\sum_{i=1}^{N_f}
		||f_{\theta}(x_i) - f_i||_2^2 \\ +
	 \frac{w_b}{N_b}\sum_{i=1}^{N_b}
		||b_{\theta}(x_i)-b_i||_2^2
	+  \frac{w_u}{N_u}\sum_{i=1}^{N_u}
		||u_{\theta}(x_i) - u_i||_2^2
	+ \frac{w_{\lambda}}{N_{\lambda}}\sum_{i=1}^{N_{\lambda}}
		||\lambda_{\theta}(x_i) - \lambda_i||_2^2.
\end{multline}
In this equation, $\{w_f, w_b, w_u, w_{\lambda}\}$ are objective function weights for balancing the various terms in Eq.~\eqref{eq:intro:piml:pinn:loss:1}.
Following determination of $\hat{\theta}$, $u_{\hat{\theta}}$ and $\lambda_{\hat{\theta}}$ can be evaluated at any $x$ in $\Omega$, $\Omega_{\lambda}$, respectively.


For neural operators, $\pazocal{F}_{\lambda, \xi}$, $\pazocal{B}_{\lambda, \xi}$ are unknown and stochastic, while $\lambda(x; \xi), u(x; \xi)$ are partially unknown. Note that the stochasticity of $\pazocal{F}_{\lambda, \xi}$ and $\pazocal{B}_{\lambda, \xi}$ arises from the stochasticity in $\lambda(x;\xi)$ with $\xi \in \Xi$.
The available data consists of paired realizations of $f, b, u$, and $\lambda$, each pair associated with a different $\xi \in \Xi$.
The realizations are considered noisy in general for $u$, and clean for $f, b$, and $\lambda$.
Further, the objective of this problem is to learn the operator mapping defined by Eq.~\eqref{eq:intro:piml:pinn:pde}. The DeepONet method, developed in \cite{lu2021learning} and designed for general nonlinear operator learning, addresses this problem by constructing an approximator using NNs, denoted by $G_\theta$ and parameterized by $\theta$, to model the mapping from the input function domain to the output function domain. In the following and without loss of generality, we consider both $f$ and $b$ in Eq.~\eqref{eq:intro:piml:pinn:pde} as known and the problem becomes learning the mapping from the parameter $\lambda$ to the solution $u$. Further, each value of $\xi \in \Xi$ in Eq.~\eqref{eq:intro:piml:pinn:pde} corresponds to a different $\lambda(x; \xi)$ function, and thus to a different solution $u(x; \xi)$.
For any $\xi\in\Xi$, the approximator $G_\theta$ takes as input a representation of $\lambda(x;\xi), x\in \Omega_\lambda$, and outputs a function, $G_\theta(\lambda)(x;\xi), x\in \Omega$. The DeepONet consists of two sub-networks, one that takes the coordinates $x\in\Omega$ as input, called trunk net, and one that takes the discretized function $\lambda$ as input, called branch net.
Considering a one-dimensional $u$, i.e., $D_u=1$, each of the two networks produces a $w$-dimensional output.
The two outputs are merged by computing their inner product for producing the final prediction $G_\theta(\lambda)(x), x\in\Omega$.
For learning the mapping from $\lambda$ to $u$ using the vanilla DeepONet \cite{lu2021comprehensive}, the minimization problem with the mean squared error (MSE) loss
\begin{equation}\label{eq:intro:piml:don:min}
	\hat{\theta} =  \underset{\theta}{\mathrm{argmin}}~\pazocal{L}(\theta) \text{, where } 
	\pazocal{L}(\theta) = \frac{1}{NN_u}\sum_{i=1}^{N}\sum_{j=1}^{N_u}
	||G_{\theta}(\lambda(.; \xi_i))(x_j) - u_j^{(i)}||_2^2    
\end{equation}
is solved, using the training dataset $\cD' = \{x_j, \xi_i, \lambda^{(i)}, u_j^{(i)}\}_{i, j= 1, 1}^{N, N_u}$, where $\{x_j\}_{j=1}^{N_u}$ are the locations, $\{\xi_i\}_{i=1}^N$ are the sampled events and in this case represent the unique pairing between $u$ and $\lambda$, $u_j^{(i)} := u(x_j, \xi_i)$ is the measurement of $u$ at the location $x_j$ for event $\xi_i$, and $\lambda^{(i)}$ is the representation of $\lambda(x; \xi_i)$. Note that, to be consistent with neural ODEs/PDEs, we use $u_\theta$ to denote the approximator for operators in the rest of this paper. In this case, $u_\theta = u_\theta(x;\lambda(\cdot;\xi))$. DeepONets, once trained, could act as PDE-agnostic surrogates encoding the hidden physics for further learning and inferences \cite{meng2022learning}.
Finally, note that in addition to supervised learning using data as described above, DeepONet can alternatively be trained by employing only the PINN loss during training.
As shown in \cite{goswami2021physicsinformed}, the most effective training is the hybrid one, where both physics and data are used in the training loss; see also \cite{wang2021learning}. 

\subsection{Uncertainty quantification}\label{sec:uqINsciml:uq}

In contrast to the point estimates obtained via the optimization problems of Eqs.~\eqref{eq:intro:piml:pinn:loss:1} and \eqref{eq:intro:piml:don:min}, the goal of UQ is to estimate the predictive distribution
\begin{align}\label{eq:pred_dist}
    p(u|x,\cD) = \int p(u|x, \theta) p(\theta | \cD) d\theta,
\end{align}
which is often referred to as Bayesian model average (BMA).
In Eq.~\eqref{eq:pred_dist}, $p(\theta|\cD)$ denotes the \textit{posterior density}, which can be obtained using Bayes' rule via
\begin{equation}\label{eq:uqt:pre:bma:post}
	p(\theta|\cD) =\frac{p(\cD|\theta)p(\theta)}{p(\cD)}.
\end{equation}
In Eq.~\eqref{eq:uqt:pre:bma:post}, $p(\cD|\theta)$ is the likelihood of the data, i.e., $p(\cD|\theta) = \prod_{i=1}^{N}p(u_i|x_i, \theta)$ for independent and identically distributed (i.i.d.) data; $p(\theta)$ is the probability density function for the prior distribution of the parameters $\theta$; and $p(\cD)$ is called \textit{marginal likelihood} or \textit{evidence}, because it represents the probability that we observe $\cD$.
The evidence is given as
\begin{equation}\label{eq:uqt:pre:bma:marglike}
	p(\cD) =\int p(\cD|\theta)p(\theta) d\theta. 
\end{equation}
The uncertainty of parameters $\theta$ is typically referred to as \textit{epistemic uncertainty}.

\begin{figure}[h]
	\centering
	\includegraphics[width=.8\linewidth]{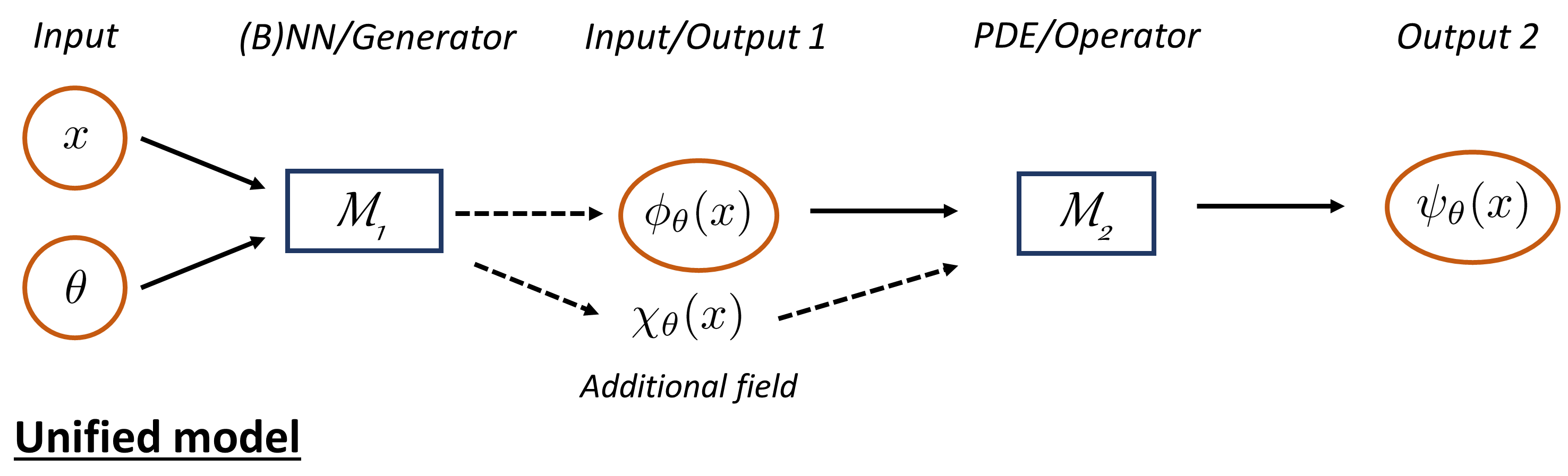}
	\caption{A unified view of the UQ models studied in Section~\ref{sec:uqINsciml:uq}; adopted from \cite{psaros2022uncertainty}. 
	The model $\pazocal{M}_1$ can be a NN or a generator network, or nothing.
	The model $\pazocal{M}_2$ can be a PDE or an operator, such as a pre-trained DeepONet.
	The quantities $\phi$, $\psi$ can be the solution $u$ and the source $f$ in Eq.~\eqref{eq:intro:piml:pinn:pde}, or vice-versa; $x$ represents the space-time input domain; and $\theta$ represents the parameters of the NN or the input to the generator.
	Further, ``$\chi$'' is used for modeling additional fields and can represent, for example, the model parameters $\lambda$ in Eq.~\eqref{eq:intro:piml:pinn:pde}. The dashed arrow indicates that the connection may not exist for some problem scenarios.} 
	\label{fig:neurals}
\end{figure}

Generally, inferring $\theta$ given $\cD$, i.e., computing the posterior exactly via Eq.~\eqref{eq:uqt:pre:bma:post} is computationally and analytically intractable. 
This is addressed by approximate posterior inference, which approximates the posterior by another distribution and/or obtains representative samples from the posterior. Specifically, the Bayesian and the deterministic frameworks are the two main families of UQ methods for SciML. The former employs either a sampling method or a variational inference (VI) method to obtain samples from the posterior distribution, whereas the latter can be interpreted as an extension of standard training of deterministic NNs. Various UQ methods have been developed for SciML within these two frameworks. 
We adopt the unified framework proposed in \cite{psaros2022uncertainty} to include all the UQ models employed in this study in Fig.~\ref{fig:neurals}, and present a short overview for them in the rest of this section.

For neural ODEs/PDEs, the solution $u$ and/or the problem parameters $\lambda$ are modeled with \textit{surrogates} parameterized by $\theta$ and denoted by $\cM_1$ in Fig.~\ref{fig:neurals}. 
The Bayesian PINNs (B-PINNs) \cite{yang2020bayesian} and the functional prior (FP) based on the physics-informed generative adversarial networks (PI-GANs) \cite{meng2022learning,yang2020physicsinformed} are two recently proposed  physics-informed learning approaches with UQ within the Bayesian framework. The Bayesian neural networks (BNNs) and the generators from trained PI-GANs are employed as the surrogate for $u/\lambda$ in these two methods, respectively. Further, either sampling or variational inference (VI) approaches can be utilized for obtaining the posterior samples for $\theta$. In addition to the aforementioned methods, the Monte Carlo dropout \cite{gal2016dropout} (MCD) as well as the deep ensemble \cite{lakshminarayanan2017simple} (DEns) approaches, which employ standard, i.e., deterministic, NNs as surrogates for $u$ and/or $\lambda$, can be combined with PINNs to quantify uncertainties in problems with known physics \cite{zhang2019quantifying,psaros2022uncertainty}. 

For neural operators, a DeepONet parameterized by $\theta$ is employed as the surrogate for the solution $u$ and denoted by $\cM_2$ in Fig.~\ref{fig:neurals}. 
In the Bayesian framework, \cite{lin2021accelerated} employed BNNs in DeepONet and a sampling method is then utilized to obtain the posterior samples for the parameters in BNNs. As for deterministic methods, extensions of standard training of DeepONets, such as DEns, are used for quantifying uncertainties in predictions \cite{psaros2022uncertainty,yang2022scalable,pickering2022discovering}. 
These models correspond to Fig.~\ref{fig:neurals} with $\cM_2$ being a DeepONet and the dashed arrows are not used. Further, the inference input in these models, i.e. the input to the branch net, should be represented in the same way as used in the training, i.e., the number of the discrete values as well as their locations should be the same. For a more general case in which the inference input and/or output data are noisy and measured at a certain number of random locations, \cite{meng2022learning} proposed to combine a pre-trained DeepONet with a GAN-based FP to quantify the uncertainties in predictions. In \cite{psaros2022uncertainty}, a BNN is utilized as an alternative to the generator of a GAN in \cite{meng2022learning}.
These two models correspond to Fig.~\ref{fig:neurals} with $\cM_1$ being either a BNN or the generator from a pre-trained GAN and $\cM_2$ a pre-trained DeepONet.

In all cases, different $\theta$ values correspond to different functions $u_{\theta}$. We refer to the surrogate combined with a distribution $p(\theta)$ for $\theta$  as a \textit{process}. Following approximate posterior inference, the posterior samples for $\theta$ are used for obtaining the statistics of the function samples.
In this regard, the value of $u$ at a location $x$ given the data $\cD$ is a random variable denoted by $(u|x, \cD)$.
The mean of $(u|x, \cD)$ is denoted by $\hat{u}(x) = \mathbb{E}[u|x, \cD]$ and approximated by $\bar{\mu}(x)$ as
\begin{equation}\label{eq:uqt:pre:mcestmc:mean}
	\hat{u}(x) \approx \bar{\mu}(x) = \frac{1}{M}\sum_{j=1}^Mu_{\hat{\theta}_j}(x),
\end{equation}
where $\{u_{\hat{\theta}_j}(x)\}_{j=1}^M$ is the set of NN predictions corresponding to the samples $\{\hat{\theta}_j\}_{j=1}^M$.
Further, the diagonal part of the covariance matrix of $(u|x, \cD)$ is given by
\begin{equation}\label{eq:uqt:pre:mcestmc:totvar}
	Var(u|x, \cD) \approx \bar{\sigma}^2(x) =  \underbrace{\Sigma_{u}^2}_{\bar{\sigma}_a^2(x)}  + \underbrace{\frac{1}{M}\sum_{j=1}^M(u_{\hat{\theta}_j}(x)-\bar{\mu}(x))^2}_{\bar{\sigma}_e^2(x)}. 
\end{equation} 
In this equation, $\bar{\sigma}^2(x)$ represents the approximate total uncertainty of $(u|x, \cD)$, while $\bar{\sigma}_a^2(x)$ and $\bar{\sigma}_e^2(x)$ denote the aleatoric and the epistemic parts of the total uncertainty, respectively.
In practice, a Gaussian (or, e.g., a Student-t) distribution is commonly fitted to the posterior samples.
For example, we denote as $\pazocal{N}(\bar{\mu}(x), \bar{\sigma}^2(x))$ the Gaussian approximation to the samples $\{u_{\hat{\theta}_j}(x)\}_{j=1}^M$ at an arbitrary location $x$, where $\bar{\mu}(x)$ and $\bar{\sigma}^2(x)$ are obtained via Eqs.~\eqref{eq:uqt:pre:mcestmc:mean} and \eqref{eq:uqt:pre:mcestmc:totvar}, respectively.

\subsection{Performance evaluation and post-training calibration}\label{sec:calibration}

Performance evaluation is necessary for model
selection, e.g., prior/architecture optimization, comparison between UQ methods, and overall quality evaluation of the UQ design procedure \cite{pickering2022structure}.
As reviewed in \cite{psaros2022uncertainty}, performance evaluation metrics include, but are not limited to, the relative $\ell_2$ error for evaluating accuracy, the mean predictive likelihood for evaluating predictive capacity, and the root mean squared calibration error for evaluating statistical consistency; i.e., how closely the predicted model matches the data-generating distribution.

Further, calibration approaches can be adopted to improve statistical consistency in UQ for SciML \cite{psaros2022uncertainty}. Various approaches have been developed for calibration, e.g., \cite{levi2019evaluating} re-weights the predicted variance optimally to calibrate the output uncertainty; \cite{kuleshov2018accurate} approximates the mis-calibration function and applies it to all predictive distributions, and thus modifies both the output mean and variance; and \cite{zelikman2020crude} fits an empirical distribution to the scaled residuals, and also calibrates both the output mean and variance. To use these approaches, an additional dataset is required after training.
This can be a left-out calibration dataset, similar to validation and test sets.

\section{Overview and details of NeuralUQ}\label{sec:library}

In this section, we first present an overview of the NeuralUQ library, and subsequently provide details regarding its design, use, and also customization for extended capabilities. 

\subsection{Library  overview}\label{sec:overview}

\begin{figure}[ht]
	\centering
	\includegraphics[width=.9\linewidth]{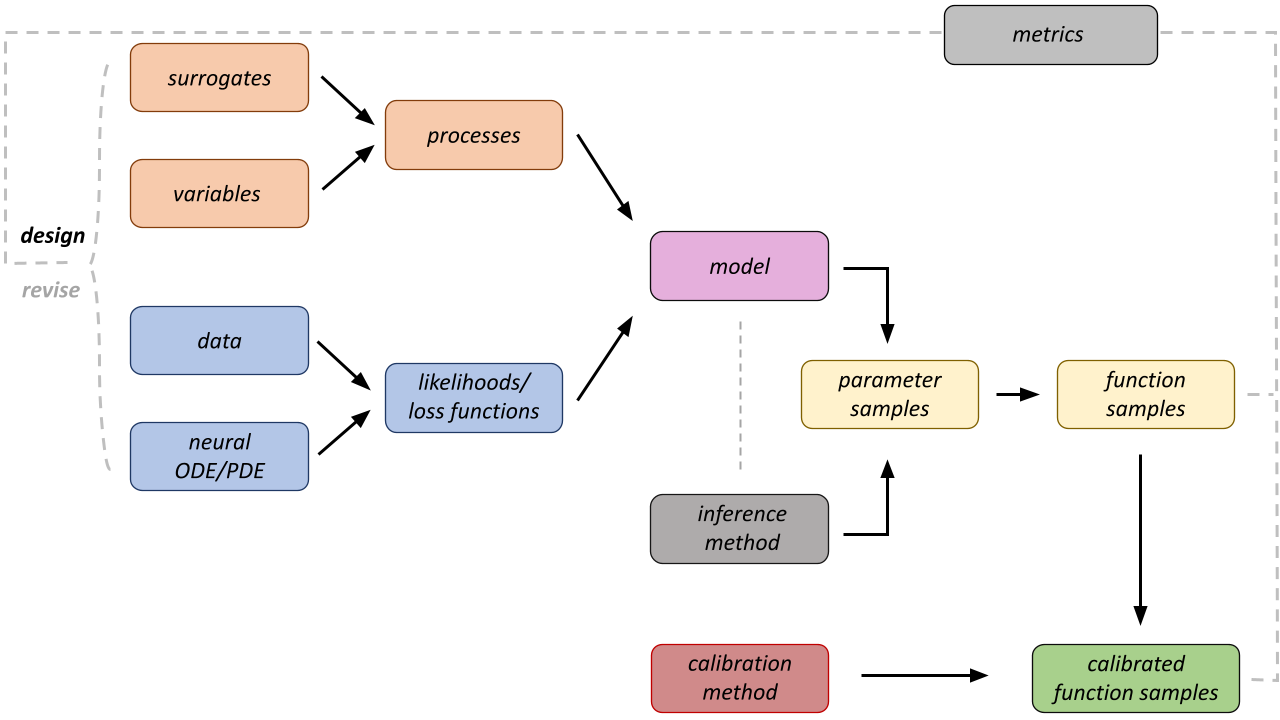}
	\caption{Library design overview.
	In the design phase, a model is constructed that includes processes representing a proposed solution family and likelihoods or loss functions representing the data distribution.
	Processes include surrogates, which can be construed as assumed functional forms, and priors distributions or variables for the parameters of the surrogates, depending on the type of method used.
    Likelihoods or loss functions are constructed using the data and the underlying neural ODE/PDE.
    To select the solution function samples that best explain the data, a posterior inference method is combined with the model.
    Finally, the obtained function samples can be calibrated using additional data. 
    In the revision phase, metrics are used for evaluating the performance and re-designing the solution procedure.
	}
	\label{fig:lib:overview}
\end{figure}

\begin{algorithm}[ht]
\caption{NeuralUQ for solving SciML problems.}
\label{alg:NeuralUQ}
\begin{enumerate}
\item[Step 1]  Specify a process, which is composed of a surrogate as well as its parameters $\theta$ to approximate the solution of an ODE/PDE, using the \textbf{\texttt{surrogates}}, \textbf{\texttt{variables}}, and \textbf{\texttt{process}} modules.
\item[Step 2] Specify the differential equations using automatic differentiation.
\item[Step 3] Construct the likelihoods or loss functions based on the equations, boundary/initial conditions, and/or given data, using the \textbf{\texttt{likelihoods}} module.
\item[Step 4] Combine the process and likelihoods/loss functions to construct a \textbf{\texttt{model}}.
\item[Step 5] Specify an inference method using the \textbf{\texttt{inference}} module, and compile the model with it to obtain a sampler for posterior distributions by calling \textbf{\texttt{model.compile}}.
\item[Step 7] Call \textbf{\texttt{model.run}} to run the compiled sampler to obtain posterior samples.
\item[Step 8] Compute predictions with uncertainties based on Eqs.~\eqref{eq:uqt:pre:mcestmc:mean} and \eqref{eq:uqt:pre:mcestmc:totvar} using the posterior samples of $\theta$ from Step 7.
\item[Step 9] Use the \textbf{\texttt{metrics}} module for performance  evaluations using the predictions from Step 8.
\item[Step 10] Use the \textbf{\texttt{calibrations}} module for post-training calibration (if applicable).
\end{enumerate}
\end{algorithm}

As mentioned in the previous section, the value of $u$ at a location $x$ given the data $\cD$ is a random variable denoted by $(u|x, \cD)$. 
The aim of UQ in the library is to obtain, calibrate, compute the statistics, and evaluate the quality of samples drawn from the posterior distribution $p(u|\pazocal{D})$.
To that end, in the design phase a model is constructed that includes surrogates representing the solution $u$ and potentially $f$ and/or $\lambda$, as well as likelihood functions representing the data distribution.
Surrogates are parameterized by $\theta$ and when combined with probability distributions of $\theta$ are represented as stochastic processes; i.e., a proposed family of solutions.
To obtain the function samples for the solutions, a posterior inference method is combined with the model. 
Finally, the obtained function samples can be calibrated using additional data. In the revision phase, metrics are used for evaluating the performance and re-designing the solution procedure. 
A schematic workflow of the library is provided in Fig.~\ref{fig:lib:overview} and Procedure~\ref{alg:NeuralUQ}.

\subsection{Library design}
\label{sec:lib:design}


\begin{table}[ht]
	\centering
	\footnotesize
	\begin{tabular}{cccc}
	\toprule
	\multicolumn{4}{c}{\textbf{UQ methods in NeuralUQ}}\\
		\midrule
		UQ framework & UQ family & Process & Inference methods    \\
		\midrule
		\multirow{2}{*}{\bf{Bayesian}} & \textit{Samplable} & BNNs, Gens &  HMC, LD, MALA, NoUTurn  \\ 
		
		
		& \textit{Variational} & BNNs, Gens & MFVI, MCD  \\ 
		\midrule
		\bf{Deterministic} & \textit{Trainable} & FNNs, Gens, DeepONets & DEns, SEns, LA\\
	\bottomrule
	\end{tabular}
	\caption{
	Summary of the UQ methods implemented in NeuralUQ.  NeuralUQ includes three different UQ method families, i.e., \textit{Samplable}, \textit{Variational}, and \textit{Trainable}. 
	Each UQ method family has various processes and inference methods, e.g.,
	BNNs+HMC in Samplable; BNNs+MFVI Variational; and FNN+DEns in Trainable. Gens: Generators in GANs.
	Short and detailed descriptions for these inference methods are provided in Table~\ref{tab:inferences} and \cite{psaros2022uncertainty, roberts1998optimal, xifara2014langevin, hoffman2014no, betancourt2017conceptual}, respectively.  Note that BNNs/Generators/FNNs combined with known ODEs/PDEs are used for UQ for neural ODE/PDE problems, while BNNs/Generators combined with pre-trained DeepONets and DeepONets using DEns are used for UQ in neural operator problems. 
	}
	\label{tab:lib:methods}
\end{table}

As discussed in Section~\ref{sec:uqINsciml:uq}, there are two general UQ frameworks for SciML, namely the Bayesian and the deterministic frameworks. 
The former relies on Bayes' theorem, and includes sampling methods, such as Monte Carlo Markov Chain (MCMC), and VI methods, such as mean-field variational inference (MFVI).
The latter relies on standard NN training, i.e., minimizing/maximizing an objective function, typically using the MSE or the maximum a posteriori (MAP) metric, between measurements and NN predictions.

The procedural differences of the aforementioned methods necessitate the development of three different UQ method families in our library, i.e.,  the \emph{Samplable},  \emph{Variational}, and \emph{Trainable} to include all the sampling-, VI- and standard-NN-based methods, respectively. Specifically, the sampling methods target directly the unnormalized posterior density function, without employing and optimizing any trainable variables.
The VI-based methods employ pre-defined distributions parameterized by trainable variables, which are optimized to approximate the true posterior distributions. 
Finally, the standard-NN-based UQ methods obtain point estimates for the parameters in the surrogates using an optimizer, rather than directly sampling from or approximating the posterior distributions. 
A summary of the currently supported UQ methods in NeuralUQ is provided in Table~\ref{tab:lib:methods}.

\subsection{Details of NeuralUQ components}\label{usage}

In this section, we describe the main components, built-in functionalities and how to use NeuralUQ.


\subsubsection{Processes}
\label{sec:processes}
\begin{figure}[ht]
    \centering
    \includegraphics[width=0.7\textwidth]{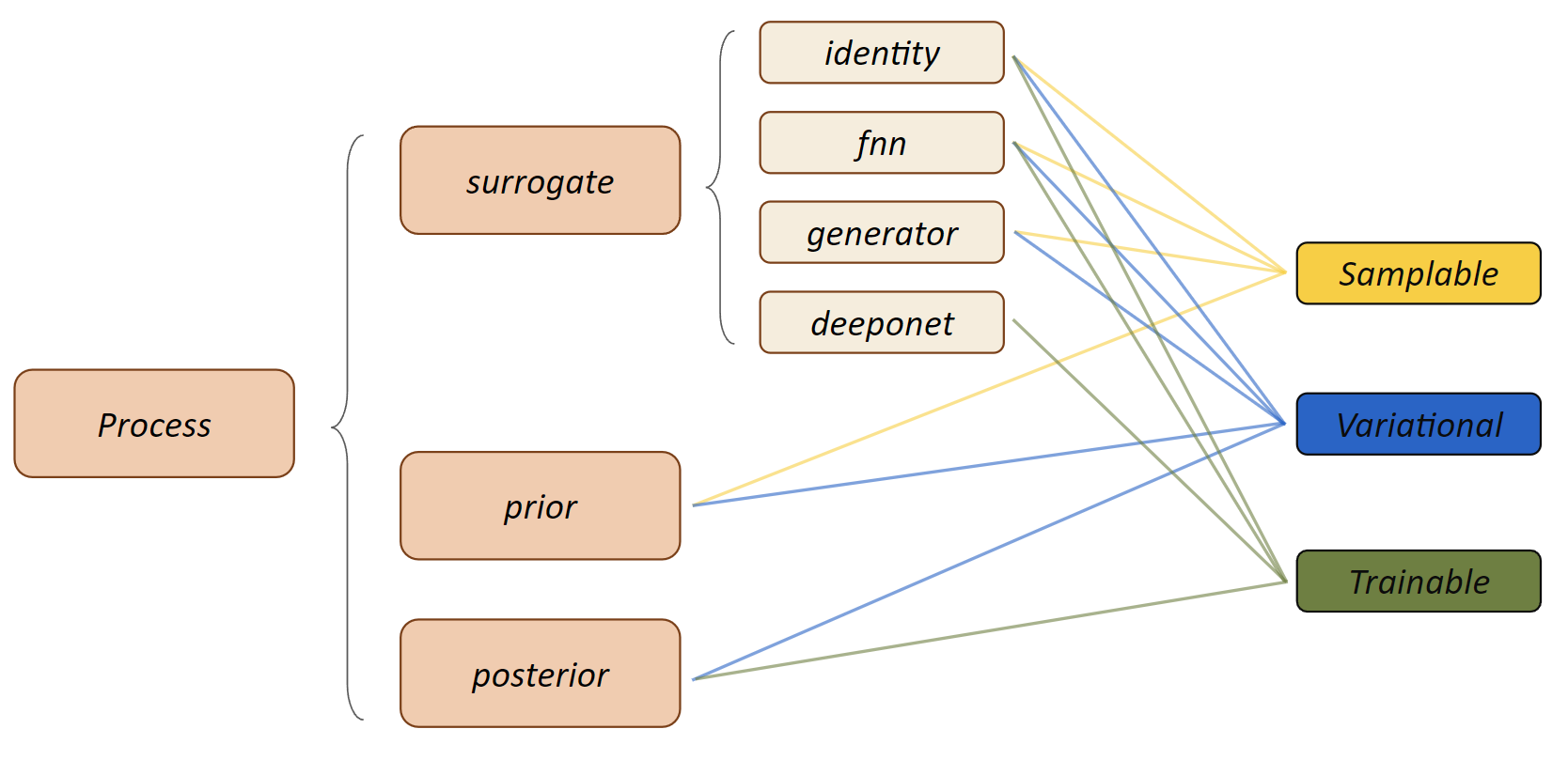}
    \caption{
    A breakdown of the \texttt{Process} module.
    A \textbf{\texttt{Process}} instance contains a surrogate as well as the attributes for its parameters. 
    Processes, used in different UQ method families, have different surrogates and attributes for their parameters, which are denoted with different line colors in this figure. Different combinations of the components yield different processes, e.g., an fnn surrogate combined with a prescribed prior distribution for its parameters results in a BNN.
    }
    \label{fig:process}
\end{figure}

As described in Procedure~\ref{alg:NeuralUQ}, the first step for using NeuralUQ is to specify a process  for the sought solution, e.g., BNNs or pre-trained generators, according to the task at hand. 
In this regard, the \textbf{\texttt{surrogates}}, \textbf{\texttt{variables}}, and   \textbf{\texttt{Process}} modules have been developed in NeuralUQ. Specifically, the \texttt{surrogates} module expresses how $x$ and $\theta$ are combined to yield the approximators $u_\theta(x)$ or $\lambda_\theta(x)$ ($\cM_1$ or $\cM_2$ in Fig.~\ref{fig:neurals}); the \texttt{variables} module is used to specify the attributes of the parameters in surrogates, e.g., prior/posterior distributions (Fig.~\ref{fig:process}); and the \texttt{Process} module is an interface for the implementation of the unified models in Fig.~\ref{fig:neurals}, which connects the surrogate with its associated parameters.


Currently, NeuralUQ provides four built-in surrogates, namely, \textbf{\texttt{surrogates.identity}}, used as a surrogate for an unknown constant; \textbf{\texttt{fnn}}, for a fully-connected neural network (FNN); \textbf{\texttt{generator}}, for the generator of a PI-GAN pre-trained with historical data; and \textbf{\texttt{deeponet}}, for a DeepONet. The \texttt{variables} module, on the other hand, is used to specify the prior and/or posterior distributions of surrogate parameters. For instance, the prior distribution is required for surrogate parameters in the Samplable family, while both the prior and the posterior distributions are required for parameters in the Variational family. The built-in variables of NeuralUQ are \textbf{\texttt{variables.const}}, used for parameters of unknown constants as well as the generators of pre-trained GANs/PI-GANs; and \textbf{\texttt{fnn}}, used for FNNs and also DeepONets.  
Further, as presented in  Fig.~\ref{fig:process} (also in Table~\ref{tab:lib:methods} and Section~\ref{sec:lib:design}),  different UQ method families support different \texttt{surrogates} and are compatible with different \texttt{variables}. In NeuralUQ, \texttt{variables.fnn} and \texttt{variables.const} are decomposed into three different sub-modules, each one pertaining to a UQ method family of Table~\ref{tab:lib:methods}, i.e., \textbf{\texttt{variables.fnn/const.Samplable}}, \textbf{\texttt{Variational}}, and \textbf{\texttt{Trainable}}. 

 With the specified surrogate and parameters, we can then instantiate a Process object to create a corresponding process. The surrogate and the attributes for its parameters can then be accessed via \textbf{\texttt{Process.surrogate}} and \textbf{\texttt{Process.prior/posterior}}, respectively.  Also, for problems with unknown parameters or multi-physics, more than one process may be employed. The module \textbf{\texttt{Process.key}} can be used to trace each process in NeuralUQ.

\subsubsection{Likelihoods} 

In the Samplable and Variational families, likelihood distributions for the given data are required to construct the unnormalized posterior distributions of the surrogate parameters.
Similarly, a metric, namely, MSE or MAP, is required for methods in the Trainable family.  
In NeuralUQ, the \textbf{\texttt{likelihoods}} module can be used to define the likelihood distributions in Samplable and Variational, as well as the MSE/MAP metrics in Trainable (i.e., Step 3 in Procedure~\ref{alg:NeuralUQ}). 
In addition, the currently supported likelihood for Samplable and Variational is the Gaussian distribution, accessed via \textbf{\texttt{likelihoods.Normal}}, and the supported metric for Trainable is the MSE, accessed via \textbf{\texttt{likelihoods.MSE}}. 
Each likelihood receives \textit{inputs}, \textit{targets} and \textit{processes} as input arguments, for specifying the locations of the measurements, the measurements, and the corresponding processes used to model them. 
For tasks involving PDEs, the argument {\textit{pde}} to the likelihood, which is a Python callable defined separately (i.e., Step 2 in Procedure~\ref{alg:NeuralUQ}), is also required.


\subsubsection{Inferences}\label{sec:inferences}

Uncertainty in the sought solution and/or unknown problem parameters is associated with the uncertainty of the surrogate parameters.
The \textbf{\texttt{inferences}} module has been developed to obtain the posterior samples for the surrogate parameters in NeuralUQ (i.e., Step 5 in Procedure~\ref{alg:NeuralUQ}).
As shown in Table~\ref{tab:lib:methods}, NeuralUQ provides various built-in inference methods in each of the UQ method families.
Specifically, NeuralUQ currently supports \textbf{\texttt{inferences.HMC}}, \textbf{\texttt{LD}}, \textbf{\texttt{MALA}}, and \textbf{\texttt{NoUTurn}} in {Samplable};  \textbf{\texttt{inferences.VI}}, \textbf{\texttt{MCD}} in {Variational}; and \textbf{\texttt{inferences.DEns}} and \textbf{\texttt{SEns}} in {Trainable}. 
Short descriptions of these methods can be found in Table~\ref{tab:inferences}.

\begin{table}[h]
	\centering
	\footnotesize
	\begin{tabular}{rcl}
		\toprule
		\multicolumn{3}{c}{\textbf{Inference methods for posterior estimation} } \\
		\midrule
		Full name & Abbreviation & Short description
		\\
		\midrule
		Hamiltonian  &\multirow{2}{*}{HMC} & Simulates Hamiltonian dynamics \\
		Monte Carlo \cite{neal2011mcmc} && followed by the Metropolis-Hasting algorithm  \\
		\arrayrulecolor{black!30}\midrule
		Metropolis-adjusted  &\multirow{2}{*}{MALA} &  Simulates Langevin dynamics \\
		Langevin algorithm \cite{roberts1998optimal, xifara2014langevin} && followed by the Metropolis-Hasting algorithm  \\
		\arrayrulecolor{black!30}\midrule
		\multirow{2}{*}{Langevin dynamics \cite{neal2011mcmc, welling2011bayesian}} & \multirow{2}{*}{LD} & \multirow{2}{*}{One leap-frog step of HMC}\\ && \\
		\arrayrulecolor{black!30}\midrule

		\multirow{2}{*}{NoUTurn \cite{hoffman2014no, betancourt2017conceptual}} & \multirow{2}{*}{NUTS} &  A variant of HMC that adaptively chooses \\ && the number of steps for the leap-frog integrator\\
		\arrayrulecolor{black!30}\midrule
		Mean-field  & \multirow{2}{*}{MFVI} & Approximates $p(\theta|\cD)$ by a Gaussian variational \\variational inference \cite{blundell2015weight}&&  distribution with diagonal covariance matrix\\ 
		\arrayrulecolor{black!30}\midrule
		Monte Carlo  & \multirow{2}{*}{MCD} & Approximates $p(\theta|\cD)$ by a Bernoulli variational \\ dropout \cite{gal2016dropout} && distribution with fixed dropout rate\\
		\arrayrulecolor{black!30}\midrule
		\multirow{2}{*}{Deep ensemble \cite{lakshminarayanan2017simple}} & \multirow{2}{*}{DEns} & Standard training $M$ times independently \\ && with random initializations \\
		\arrayrulecolor{black!30}\midrule
		\multirow{2}{*}{Snapshot ensemble \cite{huang2017snapshot}} & \multirow{2}{*}{SEns} & Standard training with learning rate\\ && schedule and snapshots\\
		\arrayrulecolor{black!30}\midrule
		Laplace  & \multirow{2}{*}{LA} & Standard training followed by fitting a \\approximation \cite{tierney1986accurate, ritter2018scalable} &&  Gaussian to approximate $p(\theta|\cD)$\\ 
		\arrayrulecolor{black}
		\bottomrule
	\end{tabular}
	\caption{
	Overview of the inference  methods in NeuralUQ; adopted from \cite{psaros2022uncertainty}.
     HMC, MALA, LD, NUTS, and MFVI are inference methods corresponding to the Bayesian framework in Table~\ref{tab:lib:methods}; and DEns to the deterministic one. Note that MCD can be  interpreted and implemented in both frameworks \cite{gal2016dropout}. In the present study, we treat it as a Bayesian method.
}
\label{tab:inferences}
\end{table}

\subsubsection{Performance evaluation and post-training calibration}

Performance evaluation, model revision and post-training prediction improvement are crucial for reliably deploying UQ methods in risk-sensitive SciML settings.
In NeuralUQ, we have developed the \textbf{\texttt{metrics}} module, which currently includes two different metrics, i.e., \textbf{\texttt{metrics.RL2E/MSE}}, and \textbf{\texttt{NLL}} (i.e., Step 9 in Procedure~\ref{alg:NeuralUQ}). 
The former serves as an evaluation metric of the predicted mean and computes the relative $L_2$ error (RL2E) or MSE between the predicted mean and the reference solution, whereas the latter serves as a metric of the predicted uncertainty and computes the negative log likelihood (NLL) \cite{yao2019quality, rahaman2021uncertainty}. 

Further, we have developed the \textbf{\texttt{calibrations}} module for performing post-training calibration (i.e., Step 10 in Procedure ~\ref{alg:NeuralUQ}). Specifically, this module currently includes one sub-module, i.e., \textbf{\texttt{calibrations.var}}, for the implementation of the method proposed in \cite{levi2019evaluating}.
We present a workflow for calibration in Procedure~\ref{alg:calibration},
and {\color{blue}}{;} a demonstration example involving post-training calibration can be found in the GitHub repository of the library. We will keep updating NeuralUQ to support more metrics and calibration methods, e.g., \cite{chung2021uncertainty}, in the future.

\begin{algorithm}[h]
\caption{A workflow for calibration using NeuralUQ.}
\label{alg:calibration}
\begin{enumerate}
\item[Step 1] Obtain uncalibrated posterior function samples at the locations $x$ where new measurements are available, using the posterior samples of the parameters from Step 7 in Procedure~\ref{alg:NeuralUQ}.
\item[Step 2] Choose and apply a calibration method from the \textbf{\texttt{calibrations}} module, which takes as input the function samples from Step 1 and new measurements.
\item[Step 3] Obtain calibrated function samples.
\end{enumerate}
\end{algorithm}

\subsection{Customization}\label{customizability}

The components of NeuralUQ, including surrogates, variables, likelihoods, and inferences, are highly flexible and customizable. 
In this section, we describe how to configure the library for extending current capabilities and addressing problem scenarios beyond the ones considered in this study.

\subsubsection{Surrogates} 

As mentioned in Section~\ref{sec:processes}, NeuralUQ currently supports four surrogates, namely, \texttt{identity}, \texttt{fnn}, \texttt{generator}, and \texttt{deeponet}. Users may need to develop and incorporate in the library alternative surrogates.
In NeuralUQ, this can be achieved in a straightforward manner by following Procedure~\ref{alg:surrogates}. 
Demonstration examples involving customizing new surrogates are presented in Appendix~\ref{sec:sine}.



\begin{algorithm}[htbp]
\caption{Customization of a new surrogate, \textbf{\texttt{MySurrogate}}, which inherits from \textbf{\texttt{surrogates.Surrogate}}.}
\label{alg:surrogates}
\begin{lstlisting}
class MySurrogate(Surrogate):
    def __call__(self, inputs, var_list):
        """
        Returns the outputs of the function, defined by samples stored in `var_list`, to the input `inputs`, and also the reshaped and tiled input, for derivative computation.
        """
\end{lstlisting}
\end{algorithm}

\subsubsection{Variables} 

As mentioned in Section~\ref{sec:processes},  \texttt{variables.fnn/const} is decomposed into three sub-modules for different UQ method families, and can be customized according to the task at hand.
In this section, customization of the variables in the first two families is discussed, as the corresponding variables influence more the performance as compared to the variables in the third one.
The built-in prior distribution for Samplable and {Variational} variables is the Gaussian distribution. 
A straightforward customization in this context pertains to modifying the mean and/or the standard deviation to obtain different priors. 
Indicatively, adding a {Samplable} in NeuralUQ variable can be done by following Procedure~\ref{alg:variables}, which includes the setup of the initial values as well as the log probability density function.  Another option pertains to selecting alternative parameterized distributions for performing VI. 
Demonstration examples involving prior customization as well as selecting alternative distributions for performing VI are presented in Appendix~\ref{sec:1d_de}.

\begin{algorithm}[htbp]
\caption{Customization of a new \texttt{Samplable} variable, \textbf{\texttt{MySamplable}}, which inherits from \textbf{\texttt{variables.\_Samplable}}.}
\label{alg:variables}
\begin{lstlisting}
class MySamplable(_Samplable):
    @property
    def initial_values(self):
        """Returns initial values of the variable."""
        
    def log_prob(self, samples):
        """Returns the log probability density of `samples`."""
\end{lstlisting}
\end{algorithm}

\subsubsection{Inference methods}

As presented in Section~\ref{sec:inferences} and summarized in Table~\ref{tab:inferences}, NeuralUQ supports various inference methods for each of the UQ method families, i.e., {Samplable}, {Variational}, and {Trainable}.
However, depending on the task at hand, users may need to develop in the library new inference methods, e.g., for reducing computational cost.
Indicatively, adding a new sampling method can be done by following Procedure~\ref{alg:inferences}.
In passing, note that all sampling methods included currently in NeuralUQ are implemented using the  TensorFlow Probability MCMC python package \cite{lao2020tfp, dillon2017tensorflow}, which is computationally efficient and straightforward to use. 
For consistency reasons, we recommend that users of NeuralUQ employ this package for adding new sampling methods to the library.


\begin{algorithm}[htbp]
\caption{Customization of a new inference, \textbf{\texttt{MyInference}}, which inherits from \textbf{\texttt{inferences.Inference}}.}
\label{alg:inferences}
\begin{lstlisting}
class MyInference(Inference):
    def make_sampler(self):
        """Creates and returns a sampler, which will be used in `sampling` method."""

    def sampling(self):
        """Performs the inference and returns samples."""
\end{lstlisting}
\end{algorithm}

\section{Demonstration examples}
\label{sec:examples}

In this section, we provide four computational examples for demonstrating the applicability and capabilities of our library.
Specifically, we quantify uncertainty in both neural ODEs/PDEs and neural operators applied to the following problems: (1) a Kraichnan-Orszag system, (2) a susceptible-infected-recovered-dead (SIRD) model for COVID-19, (3) a Korteweg-de Vries problem, and (4) a 100-dimensional Darcy problem. 
In the first three cases, we obtain and present results using uncertain PINNs, i.e., B-PINNs combined with HMC and PINNs combined with DEns, as representative examples of the Bayesian and deterministic methods, respectively. 
For the 100-dimensional Darcy problem, we present results using the DeepONet combined with two physics-agnostic models, i.e., physics-agnostic GAN functional prior (PA-GAN-FP) and physics-agnostic BNN functional prior (PA-BNN-FP),  as well as DEns.
Results corresponding to additional UQ methods and details on the computations in each case, e.g., architectures of NNs and parameters used in the training, are provided in the GitHub repository of our library.

\subsection{Kraichnan-Orszag system}\label{example:ko}


The Kraichnan-Orszag model consisting of  three coupled nonlinear ODEs describes the temporal evolution of a system with several interacting inviscid  shear waves, whose initial conditions are drawn from a Gaussian distribution \cite{wan2006multi}. 
Here, we use NeuralUQ to solve the {\em inverse} Kraichnan-Orszag problem, i.e., identify the unknowns in the governing equations given sparse and noisy measurements on the solutions without the knowledge of the initial conditions. 
The governing equations are:

\begin{align}\label{eq:ko}
    \frac{dx_1}{dt} &= a x_2 x_3, ~ \frac{d x_2}{dt} = b x_1 x_3, ~ \frac{dx_3}{dt} = -(a+b) x_1 x_2, \\
    x_1(0) &= x_2(0) = 1, ~ x_3(0) = 0.5,
\end{align}
where $a = 1$ and $b = 1$ are constants. 

\begin{figure}[ht]
    \centering
    \subfigure[]{
    \includegraphics[width=0.3\textwidth]{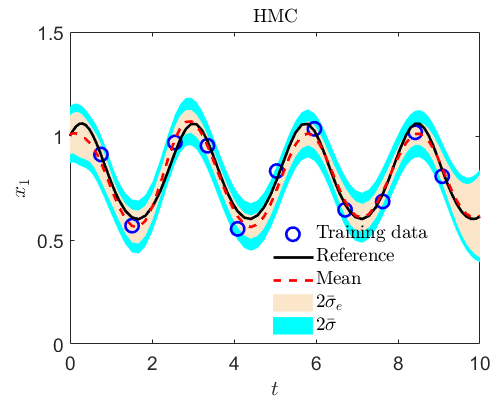}
    \includegraphics[width=0.3\textwidth]{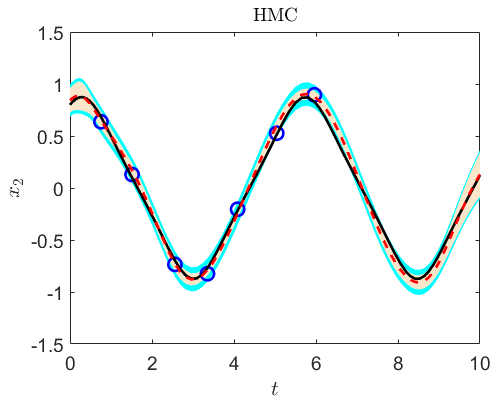}
    \includegraphics[width=0.3\textwidth]{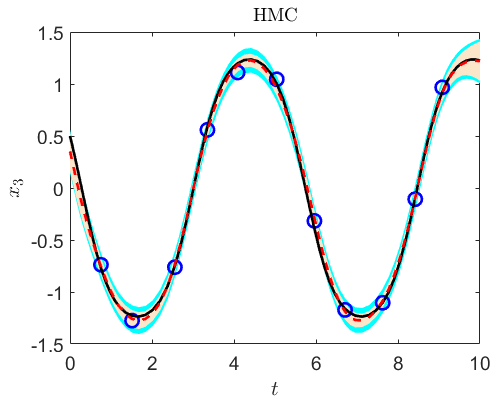}}
    \subfigure[]{
    \includegraphics[width=0.3\textwidth]{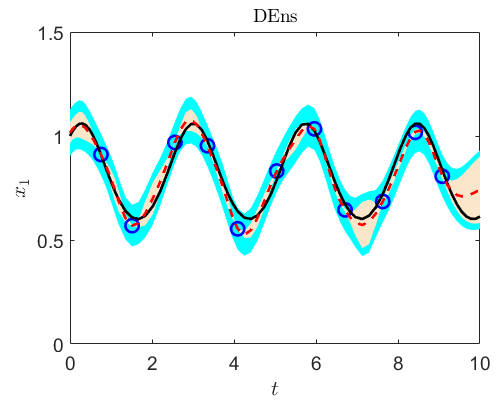}
    \includegraphics[width=0.3\textwidth]{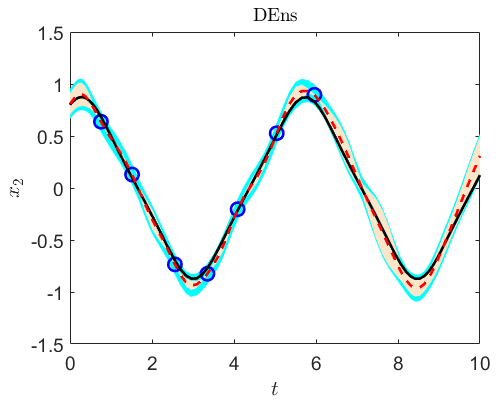}
    \includegraphics[width=0.3\textwidth]{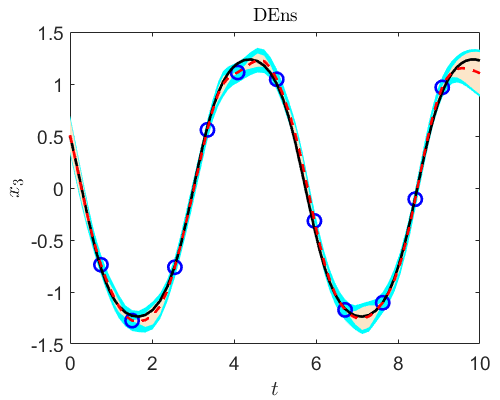}}
    \caption{
    Inverse Kraichnan-Orszag problem: Predictions corresponding to $x_1$, $x_2$, and $x_3$ using HMC (a) and DEns (b).
    Epistemic uncertainty is denoted by $2\bar{\sigma}_e$, whereas total uncertainty by $2\bar{\sigma}$.
    }
    \label{fig:ko}
\end{figure}

Particularly for this problem, we assume that $a$ and $b$ are unknown, whereas 11 noisy measurements of $x_1$ and $x_3$ are available, and 7 of $x_2$; see also Fig.~\ref{fig:ko}. 
The noise for all measurements is assumed to be Gaussian with zero mean and a known standard deviation set to 0.05. 
The objective here is to identify $a$ and $b$, as well as to reconstruct the complete solutions $x_1$, $x_2$, and $x_3$ for $t \in [0, 10]$, with uncertainties. 

We present the predictions for $x_1, x_2, x_3$ and $a, b$ using different approaches in Fig.~\ref{fig:ko} and Table~\ref{tab:ko}, respectively. 
As shown in Fig.~\ref{fig:ko}, for both HMC and DEns, the predicted uncertainty increases at locations with no available data.
Further, the error between the predicted mean and the reference solution for $x_1, x_2, x_3$ is bounded by the total uncertainty. Furthermore, as shown in Table~\ref{tab:ko}, the predicted values for $a$ and $b$, using both approaches, are close to the reference values. 
In the table, the results using MFVI and MCD are also provided for comparisons. 
In Table~\ref{tab:ko2}, we evaluate the predictions for $x_1, x_2, x_3$ using the RL2E and the NLL metrics. As shown, HMC and DEns provide the most reasonable predictions for the mean as well as uncertainties according to the computed metrics.


\begin{table}[ht]
    \footnotesize
    \centering
    \begin{tabular}{c|cccc}
    \toprule
         & HMC & DEns & MFVI & MCD  \\
         \midrule
       $a$ (mean $\pm$ std) & $0.8380 \pm 0.0813 $  & $0.8720\pm0.0812$ & $0.5706\pm 0.0205$ & $0.6141\pm 0.0202$\\
       
       $b$ (mean $\pm$ std) & $1.0705 \pm 0.0454$  & $1.1463\pm0.0558$ & $0.9388\pm 0.0166$ & $0.8931\pm 0.0161$\\
       \bottomrule
    \end{tabular}
    \caption{
    Inverse Kraichnan-Orszag problem: Predictions corresponding to $a$ and $b$ using different methods. The reference values are $a=b=1$.
    }
    \label{tab:ko}
\end{table}

\begin{table}[H]
    \footnotesize
    \centering
    \begin{tabular}{c|cccc}
    \toprule
         & HMC & DEns & MFVI & MCD \\
         \midrule
       RL2E & 0.0579 & 0.0585 & 0.2232 & 0.1527 \\
       
       NLL ($\times 10^2$) & -3.6558 & -4.0275 & -0.1187 & -1.8930 \\
       \bottomrule
    \end{tabular}
    \caption{
    Inverse Kraichnan-Orszag problem: performance evaluation of the considered methods on test dataset.
    Note that NLL is computed using the total uncertainty. Smaller RL2E and NLL mean better accuracy for the predictions of mean and uncertainty, respectively. RL2E corresponds to relative $L_2$ error, whereas NLL to the negative log likelihood. 
    }
    \label{tab:ko2}
\end{table}




\subsection{SIRD model for COVID-19}
\label{example:sird}
 Scientific models are critical tools for understanding and forecasting the spread of the COVID-19. 
 However, constructing models that are capable of accurately describing the COVID-19 still remains a challenge. 
 Here, we demonstrate the capability of NeuralUQ for learning the corresponding model with uncertainties given measurements.
 In particular, we consider the SIRD compartmental model with time-dependent parameters/coefficients/rates for the COVID-19 \cite{chen2020time, calafiore2020time, sen2021use}, expressed as:
 \begin{equation}
\begin{aligned}
    \frac{dS}{dt} &= \frac{-\beta(t) S I}{n}, ~ \frac{dI}{dt} = \frac{\beta(t) S I}{n} - \gamma(t) I - \nu(t) I,\\
    \frac{dR}{dt} &= \gamma(t) I, ~ \frac{dD}{dt} = \nu(t) I,
\end{aligned}
\end{equation}
where $\beta$, the transmission rate, $\gamma$, the recovery rate, and $\nu$, the death rate, are unknown and assumed to be time-dependent functions, and $n$ is the total population, which is assumed to be constant and equal to the summation of $S, I, R, D$ throughout the time span. $S(t), I(t), R(t)$ and $D(t)$ are the number of individuals susceptible to the disease, the number of individuals infected, the (cumulative) number of individuals recovered, and the (cumulative) number of individuals deceased from the disease, respectively, at time $t$. In this example, we use the COVID-19 dataset from Italy \cite{dong2020interactive}, which is the same as in \cite{kharazmi2021identifiability}. 


\begin{figure}[h]
    \centering
    \subfigure[]{
    \includegraphics[width=0.4\textwidth]{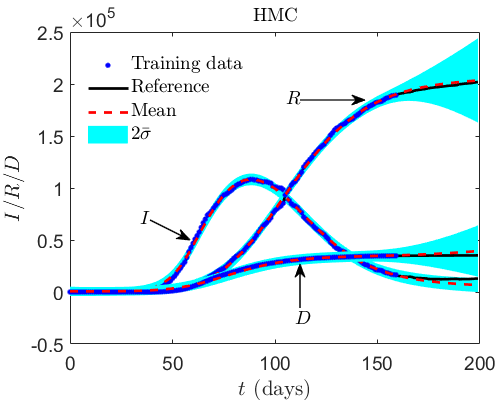}
    \includegraphics[width=0.4\textwidth]{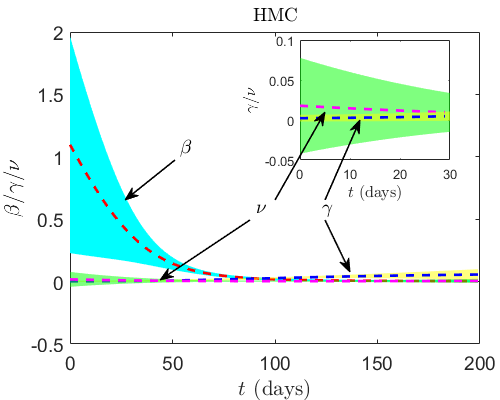}}
    \subfigure[]{
    \includegraphics[width=0.4\textwidth]{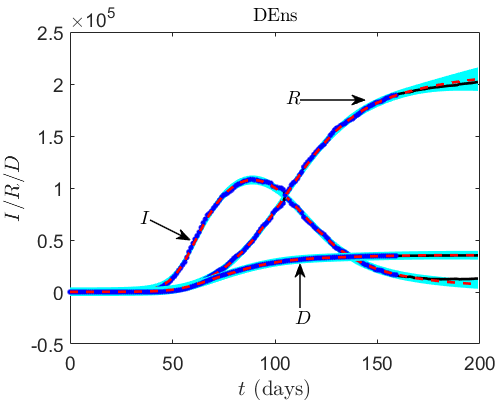}
    \includegraphics[width=0.4\textwidth]{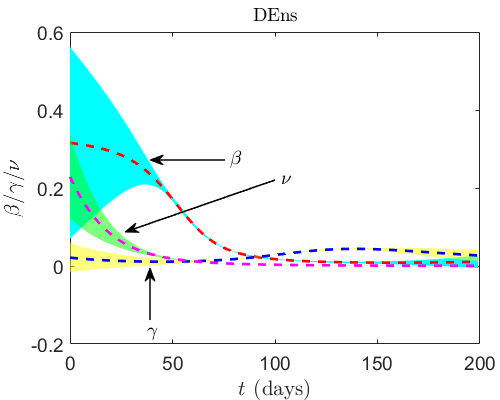}}
    \caption{
    Inverse SIRD problem: Predictions for $I$, $R$, $D$ and also $\beta$, $\gamma$ and $\nu$, using HMC (a) and DEns (b).  Total uncertainty is denoted by $2\bar{\sigma}$.
    }
    \label{fig:sird}
\end{figure}

In this problem, we utilize the data on $I$, $R$, and $D$ in the first 160 days for training.
The objective is to predict $I$, $R$, and $D$ for $t \in  [161, 200]$, and identify the unknown functions $\beta$, $\gamma$, and $\nu$ for $t \in [0, 200]$. 
The results from HMC and DEns are presented in Fig.~\ref{fig:sird}. We observe that the error between the predicted mean and the reference solution of $I/R/D$ from both approaches is bounded by the total uncertainties.
Further, HMC and DEns yield different predictions for $\beta$ and $\nu$ for times $t < 100$, while they are quite similar for $t>100$. Also, both methods predict a decrease of $\beta$ and $\nu$ with time.
In addition, the predictions for $\gamma$ from both methods are similar, i.e., they are  close to zero for $t \in [0, 200]$. 
Note that reference solutions for the unknown functions $\beta$, $\gamma$, and $\nu$, are not available and thus corresponding comparisons are not provided here.



\begin{figure}[h]
    \centering
    \subfigure[]{\label{fig:kdva}
    \includegraphics[width=0.9\textwidth]{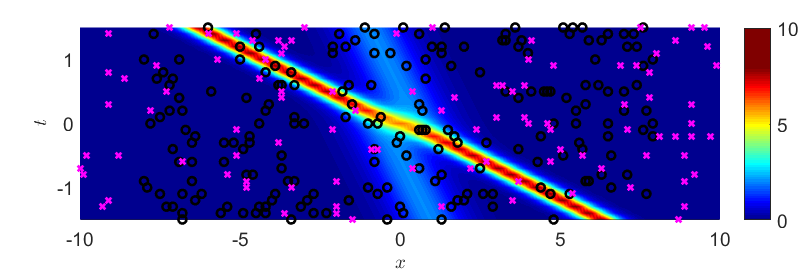}}
    \subfigure[]{\label{fig:kdvb}
    \includegraphics[width=0.4\textwidth]{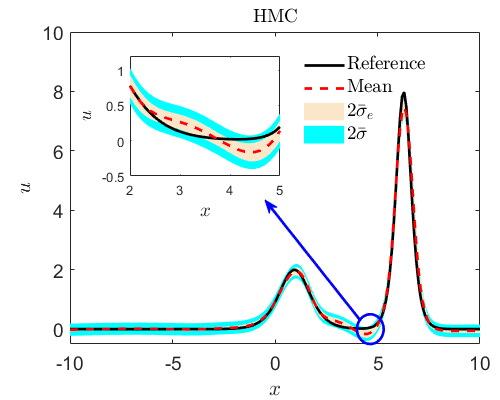}
    \includegraphics[width=0.4\textwidth]{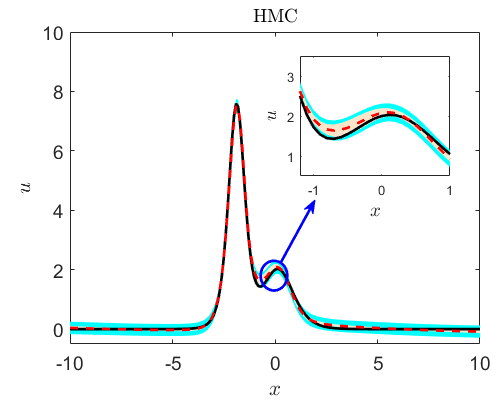}}
    \subfigure[]{\label{fig:kdvc}
    \includegraphics[width=0.4\textwidth]{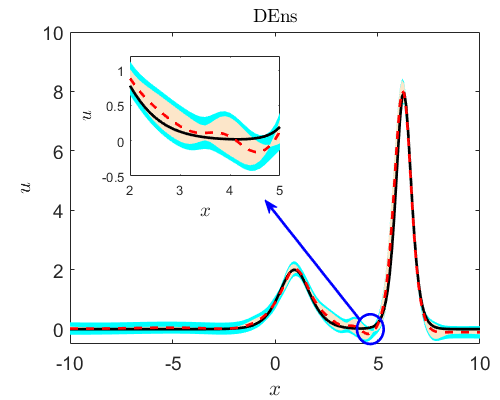}
    \includegraphics[width=0.4\textwidth]{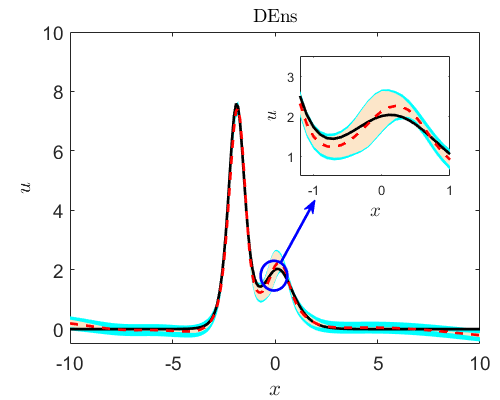}}
    \caption{
    Inverse KdV problem: Predictions for $u(x, t)$ at different times, as obtained by using HMC and DEns. (a) Locations of training data for $u$ and $f$; black circles are used for $u$ and magenta cross symbols for $f$. (b) HMC; $t = -1.5$ in the left side and $t = 0.4$ in the right side. (c) DEns; $t = -1.5$ in the left side and $t = 0.4$ in the right side. Epistemic uncertainty is denoted by  $2\bar{\sigma}_e$, whereas total uncertainty by $2\bar{\sigma}$. Note that  the  scaling for $t$ is the same as  in \cite{benes2006decompositions}. One could also use a different scaling for $t$ to make it positive only.  
    }
    \label{fig:kdv}
\end{figure}

\subsection{Inverse KdV problem}\label{example:kdv}

The Korteweg-de Vries (KdV) equation, derived more than 150 years ago, can be used to model surface waves in a shallow canal. 
It has been applied in various disciplines for modeling surface gravity waves, internal solitons in the ocean, magma flows, and conduit waves, for example. 
In this section, we solve an inverse KdV problem using NeuralUQ.
Specifically, we discover the KdV equation given data for a system with interactions between two solitary waves. 
The KdV equation is expressed in the general form:
\begin{equation}\label{eq:kdv}
u_t - \lambda_1 u u_x - \lambda_2 u_{xxx} = f,
\end{equation}
where $\lambda_1$ and $\lambda_2$ are constants, and $f = 0$ is considered. 
The exact solution can be expressed as \cite{benes2006decompositions}
\begin{equation}
    u = 2\partial^2_x R,
\end{equation}
where
\begin{equation}
    R = \log\left[\exp(-\eta_1 - \eta_2) + \exp(\eta_1 - \eta_2) + \exp(\eta_2 - \eta_1) + \frac{(a_1 - a_2)^2}{(a_1 - a_2)^2} \exp(\eta_1 + \eta_2) \right],
\end{equation}
and
\begin{equation}
\begin{aligned}
    \eta_1 &= a_1 x + a^3_1 \mu t  + b_1, ~\eta_2 = a_2 x + a^3_2 \mu t  + b_2, \\
    a_1 &= 1, ~a_2 = 2, ~b_1 = \log(3)/2, ~ b_2 = b_1, \mu = 1,
\end{aligned}
\end{equation}
where $\lambda_1 = 1.5$ and $\lambda_2 = 0.25$ are considered.

For the inverse problem, the objective is to identify $\lambda_1$ and $\lambda_2$, which are considered unknown, and to predict $u$ in the entire spatial-temporal domain, given partial noisy measurements on $u$ and $f$. 
Specifically, 200 and 100 random noisy measurements on $u$ and $f$, respectively, are considered available; see also Fig.~\ref{fig:kdva}. 
The noise for the measurements on both $u$ and $f$ is assumed to be Gaussian with zero mean and known standard deviation set to 0.1.

Shown in Figs.~\ref{fig:kdvb}-\ref{fig:kdvc} are the errors between the HMC and DEns predicted means, for $u$ at $t = -1.5$ and 0.4, and the reference solutions bounded by the predicted total uncertainties. 
Further, the predicted values for $\lambda_1$ and $\lambda_2$, which are provided in Table~\ref{tab:kdv}, are close to the reference values for both methods. 

\begin{table}[h]
    \footnotesize
    \centering
    \begin{tabular}{c|cc}
    \toprule
         & HMC & DEns  \\
         \midrule
       $\lambda_1$ (mean $\pm$ std) & 1.4095 $\pm$ 0.0372 & 1.5452 $\pm$ 0.0672  \\
       $\lambda_2$ (mean $\pm$ std) & 0.2642 $\pm$ 0.0057 & 0.3249 $\pm$ 0.0234 \\
       \bottomrule
    \end{tabular}
    \caption{
    Inverse KdV problem: Predictions corresponding to $\lambda_1$ and $\lambda_2$ using HMC and DEns; the reference values are $\lambda_1=1.5$ and $\lambda_2=0.25$.
    }
    \label{tab:kdv}
\end{table}




\subsection{100-dimensional Darcy problem}\label{example:darcy}

Flow through porous media is studied in various disciplines, such as chemical engineering, e.g., for flows in packed-bed catalytic reactors, oil recovery, e.g., for the displacement of oil with water, and in geophysics e.g., for the transport of pollutant in soil. A steady flow through porous media in two dimensions is considered here, which is described by the Darcy's law as follows:
\begin{align}\label{eq:darcy_eq}
\nabla \cdot (\lambda(x, y; \xi) \nabla u(x, y; \xi)) = f,~ x, y \in [0, 1],
\end{align}
where $\lambda$ is the hydraulic conductivity field, $u$ denotes the hydraulic head, and $f = 50$ is considered. The boundary conditions are:
\begin{subequations}\label{eq:darcy_bcs}
\begin{align}
    u(x=0, y) = 1, ~ u(x=1, y) = 0, \\
    \partial_{\boldsymbol{n}} u(x, y=0) = \partial_{\boldsymbol{n}} u(x, y=1) = 0,
\end{align}
\end{subequations}
where $\boldsymbol{n}$ denotes the unit normal vector of the boundary.

In general, $\lambda(x, y; \xi)$ is determined by the pore structure. 
In this study, we employ a stochastic model for $\lambda(x, y; \xi)$ to consider  porous media with different micro-structures. 
In particular, $\lambda(x, y; \xi) = \exp(\tilde{\lambda}(x, y; \xi))$, where $\tilde{\lambda}(x, y; \xi)$ is a truncated Karhunen-Lo\`{e}ve expansion of a Gaussian process with zero mean and the following kernel:
\begin{subequations}\label{eq:comp:don:gp}
\begin{align}
k_{\tilde{\lambda}}(x, y, x', y') = \exp(-\frac{(x - x')^2}{2l^2} - \frac{(y - y')^2}{2l^2}), \\
    0 \le x, y, x', y' \le 1, ~ l = 0.25,
\end{align}
\end{subequations}
and we only keep the first 100 leading terms of the expansion. 

We consider two different scenarios here, i.e., Case A, in which we employ a DeepONet mapping $\tilde{\lambda}$ to $u$ pre-trained with 9900 different paired data  ($\tilde{\lambda}$, $u$), and the test data are partial noisy measurements on $\lambda$ and $u$; and Case B, in which we employ the DEns method to train a DeepONet using the same training data as in Case A, which also maps $\tilde{\lambda}$ to $u$. 
Complete $\lambda$ measurements are used as test data, i.e., a measurement is available at each required location of the DeepONet input. Particularly, we refer to the DeepONet trained with DEns as uncertain DeepONet (U-DeepONet) following \cite{psaros2022uncertainty}. 


\subsubsection{Physics-agnostic functional prior}\label{example:fp}
\begin{figure}[ht]
    \centering
    \subfigure[]{
    \includegraphics[width=0.3\textwidth]{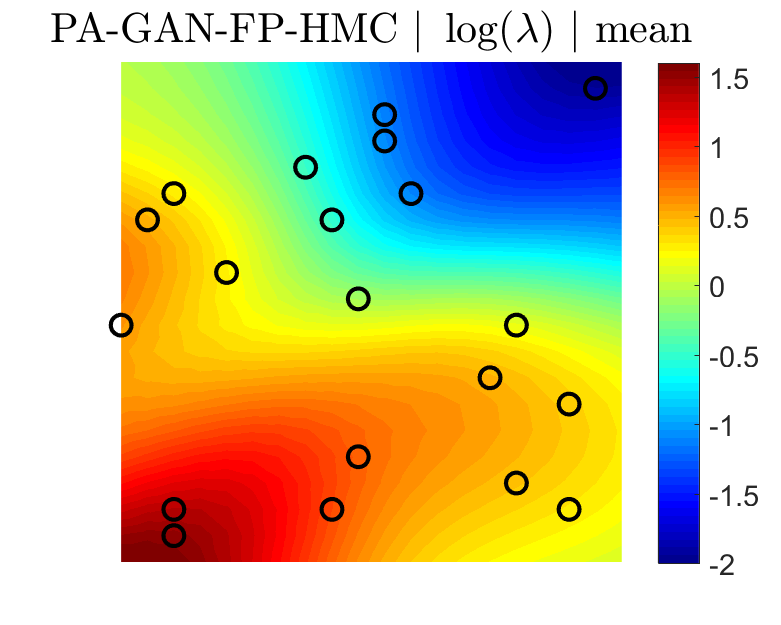}
    \includegraphics[width=0.3\textwidth]{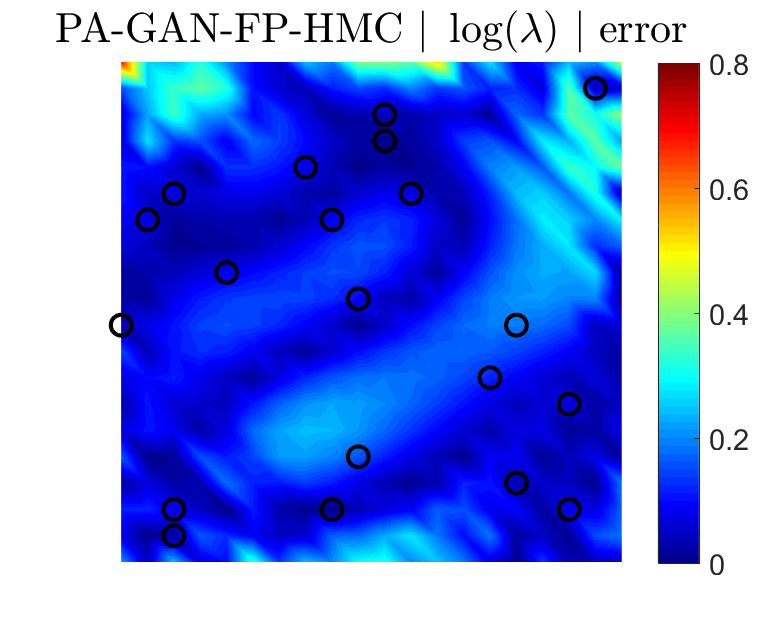}
    \includegraphics[width=0.3\textwidth]{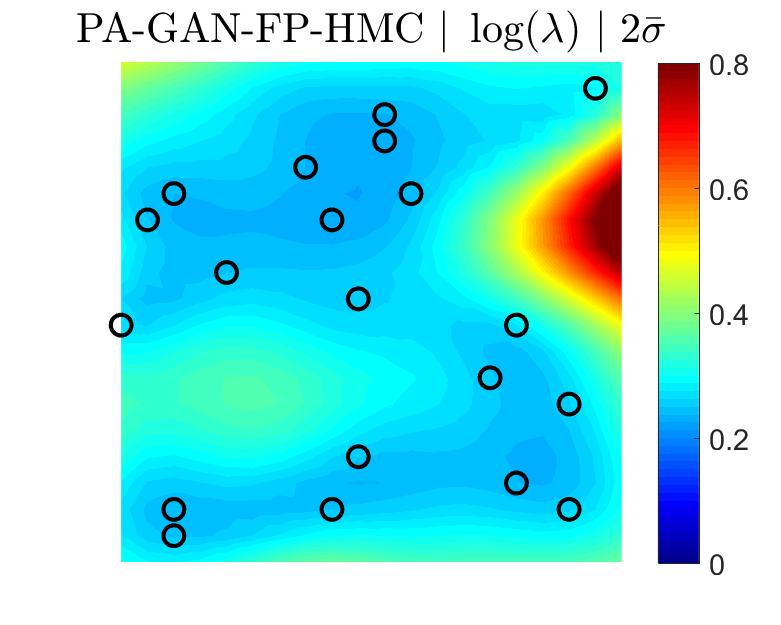}
    }
    \subfigure[]{
    \includegraphics[width=0.3\textwidth]{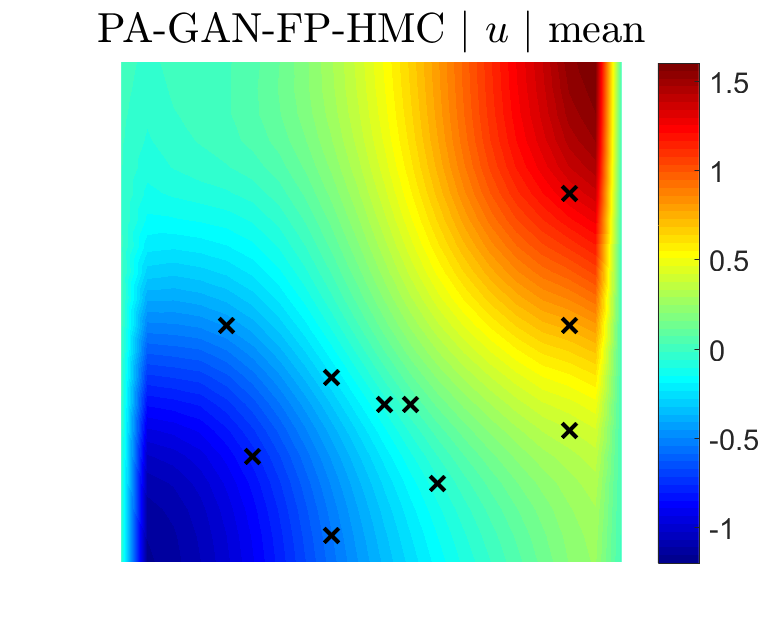}
    \includegraphics[width=0.3\textwidth]{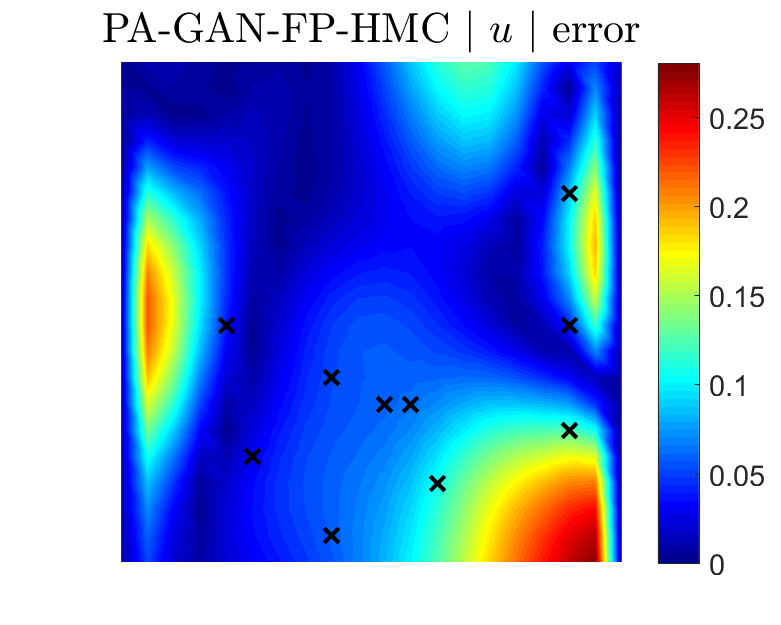}
    \includegraphics[width=0.3\textwidth]{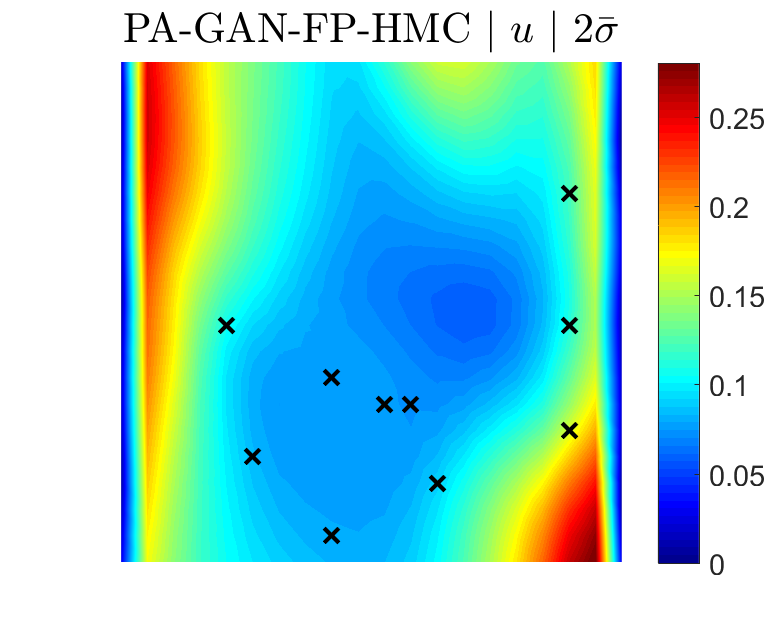}
    }
    \caption{
    100-dimensional Darcy problem (Case A): Predictions for $\log(\lambda)$ and $u$ from PA-GAN-FP+HMC. Training data for $\lambda$ are shown with circles, whereas training data for $u$ with cross symbols.
    Total uncertainty is denoted by $2\bar{\sigma}$.
    }
    \label{fig:darcy_fp_hmc}
\end{figure}

\begin{figure}[h]
    \centering
    \subfigure[]{
    \includegraphics[width=0.3\textwidth]{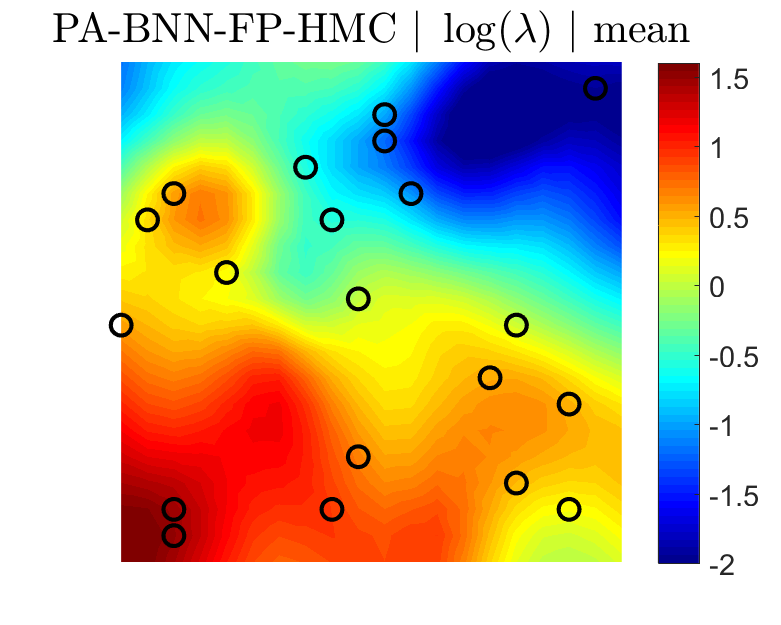}
    \includegraphics[width=0.3\textwidth]{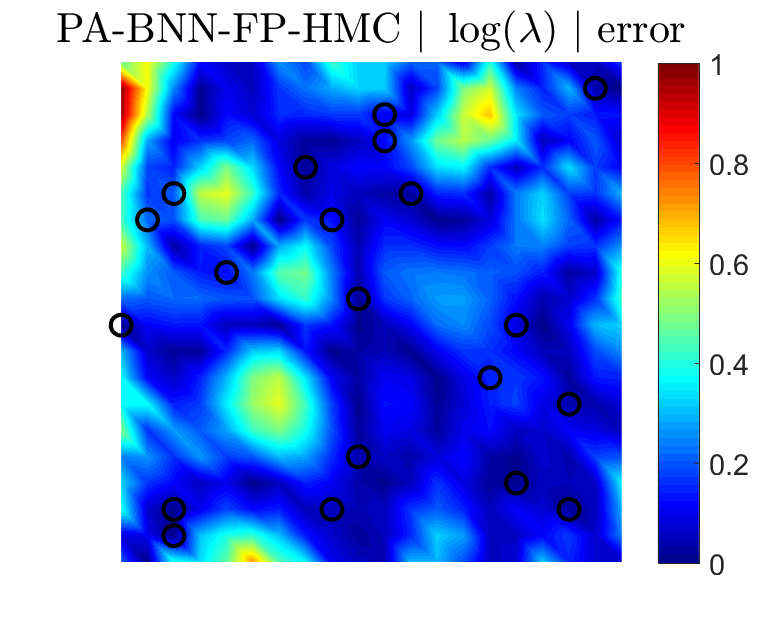}
    \includegraphics[width=0.3\textwidth]{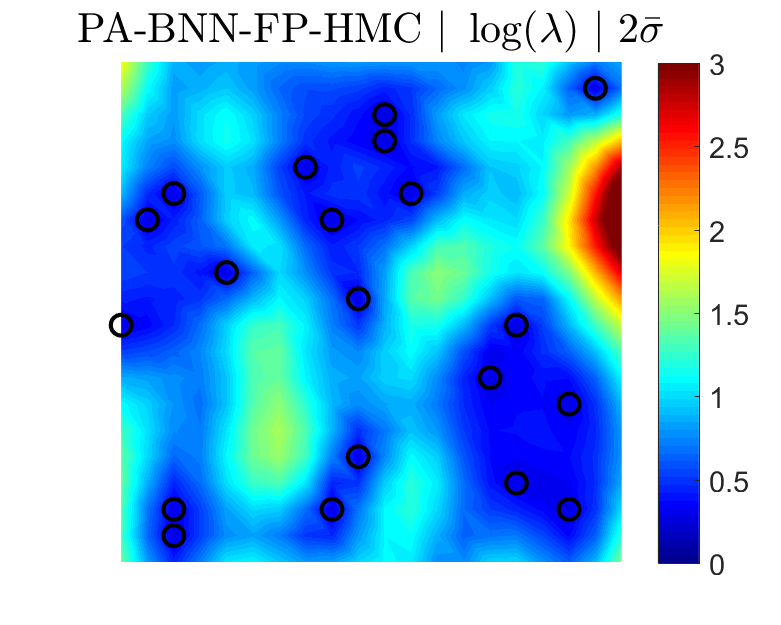}
    }
    \subfigure[]{
    \includegraphics[width=0.3\textwidth]{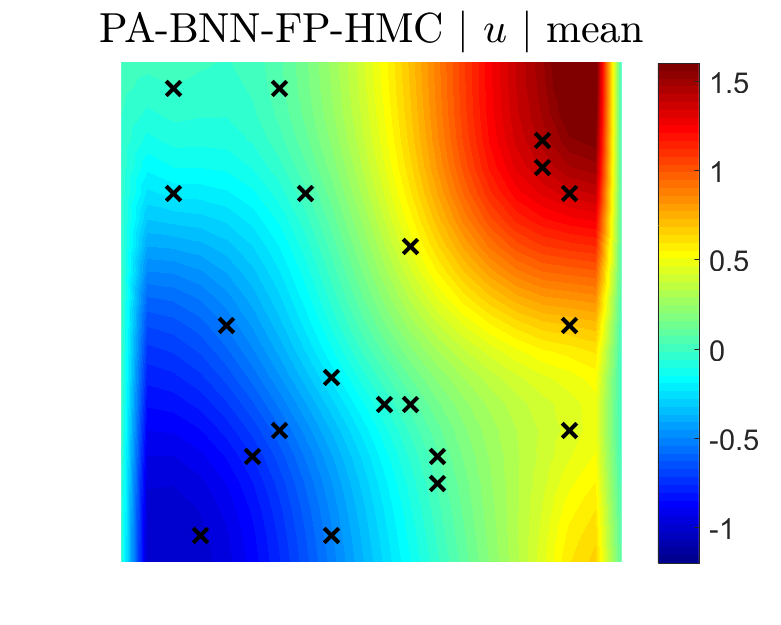}
    \includegraphics[width=0.3\textwidth]{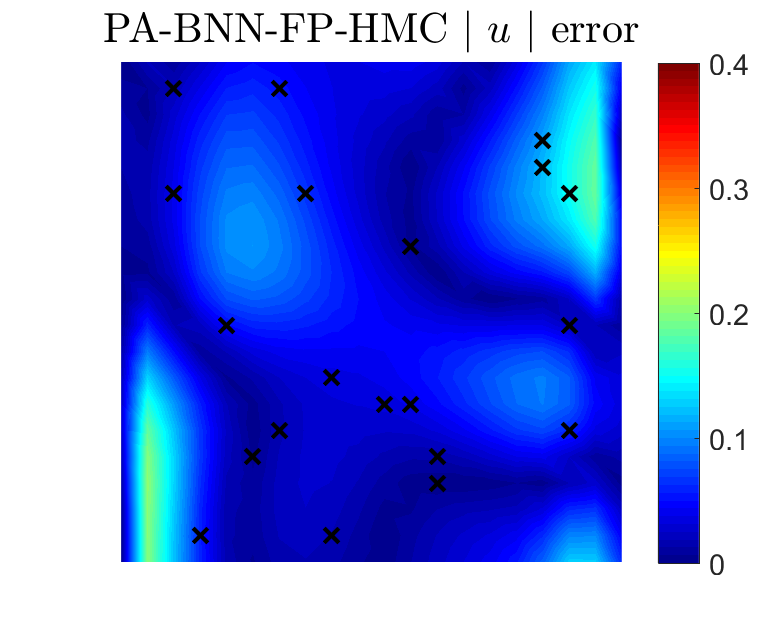}
    \includegraphics[width=0.3\textwidth]{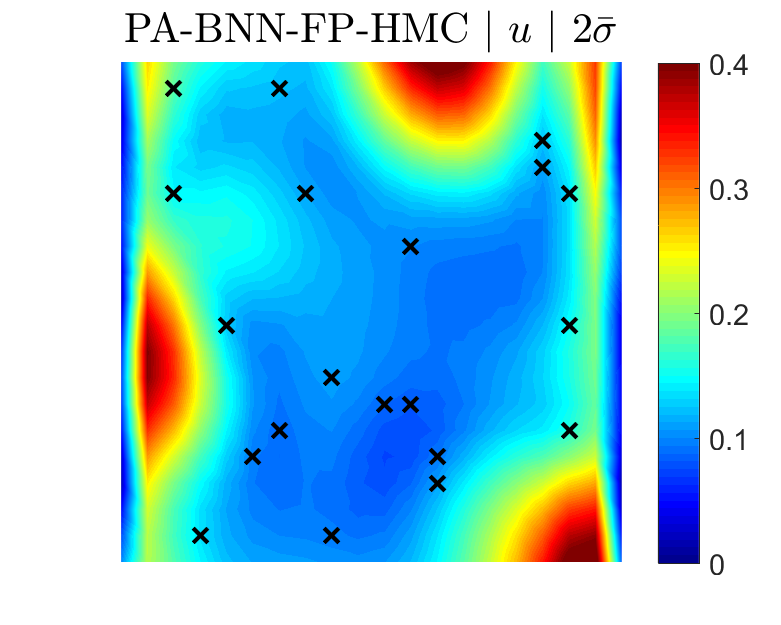}
    }
    \caption{
    100-dimensional Darcy problem (Case A): Predictions for $\log(\lambda)$ and $u$ from PA-BNN-FP+HMC. Training data for $\lambda$ are shown with circles, whereas training data for $u$ with cross symbols.
    Total uncertainty is denoted by $2\bar{\sigma}$.
    }
    \label{fig:darcy_bnn_hmc}
\end{figure}

For Case A, 20 noisy measurements of $\lambda$ and $u$ are assumed to be available; see also Figs.~\ref{fig:darcy_fp_hmc} and \ref{fig:darcy_bnn_hmc} for the respective locations. 
The objective is to reconstruct the entire $\lambda$ and $u$ given partial noisy measurements of $\lambda/u$. Specifically, a pre-trained DeepONet combined with two physics-agnostic models, i.e. a PA-BNN-FP and a PA-GAN-FP, are considered. In particular, the PA-GAN-FP is pre-trained using the same data for training the DeepONet to learn the functional prior for $\tilde{\lambda}$, while we specify a standard normal distribution as the prior for each parameter in the PA-BNN-FP for the prior of $\tilde{\lambda}$.  For posterior inference, HMC is used for both models. 
The details for the pre-trained of DeepONet and PA-GAN-FP can be found in the GitHub repository.

With the pre-trained DeepONet and the physics-agnostic models as the prior in function space, we present the predictions for $\lambda$ and $u$ in Figs.~\ref{fig:darcy_fp_hmc} and \ref{fig:darcy_bnn_hmc}, respectively.
We observe that the errors between the predicted means and the reference solutions for $\lambda$ and $u$ are mostly bounded by the predicted uncertainties in both methods.
Further, the predicted uncertainties from PA-GAN-FP are smaller than those from PA-BNN-FP, which is expected since PA-GAN-FP is equipped with a more informative prior. Interested readers can refer to \cite{meng2022learning,psaros2022uncertainty} for more comparison between the GAN- and BNN-FP.

\subsubsection{Uncertain DeepONet}

\begin{figure}[h]
    \centering
    \includegraphics[width=0.22\textwidth]{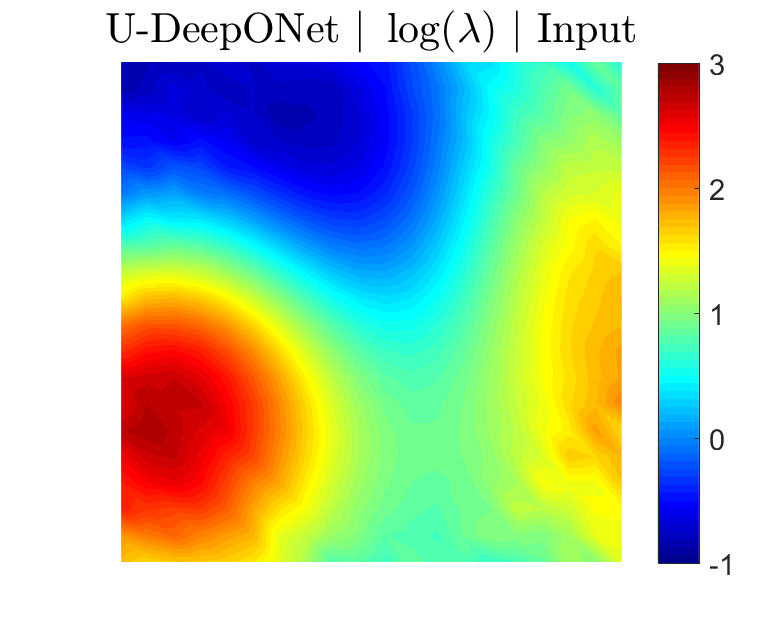}
    \includegraphics[width=0.22\textwidth]{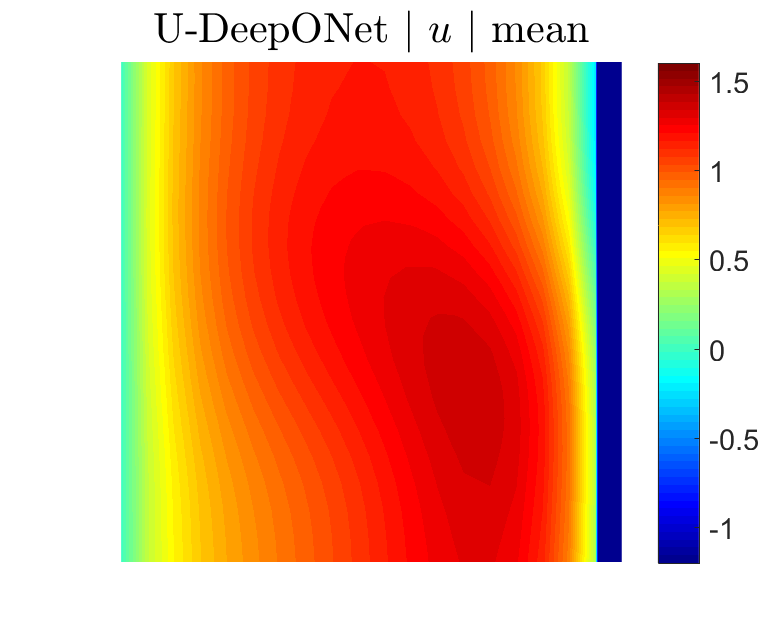}
    \includegraphics[width=0.22\textwidth]{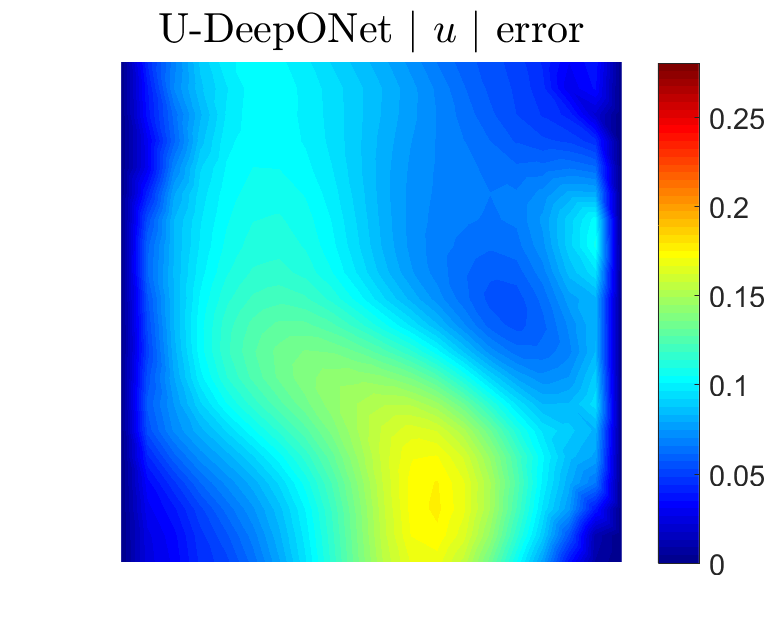}
    \includegraphics[width=0.22\textwidth]{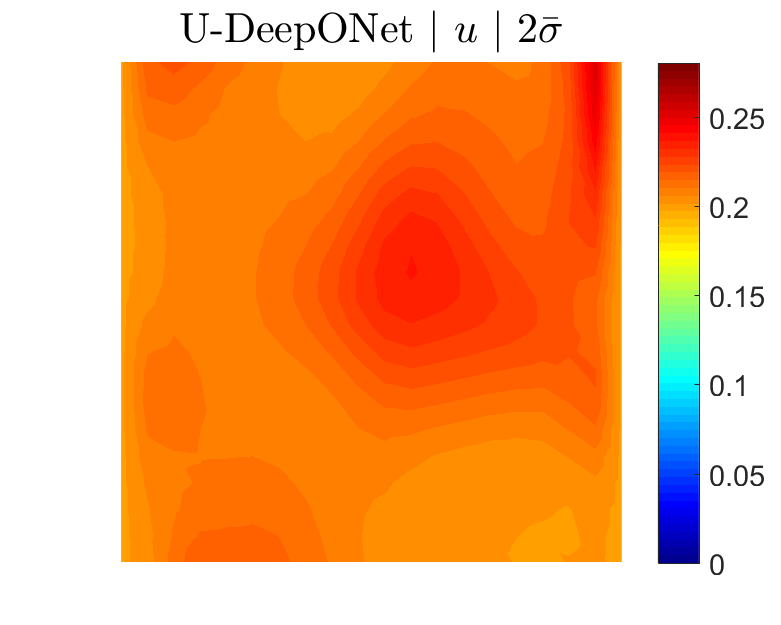}
    \caption{
    100-dimensional Darcy problem: Predictions obtained from U-DeepONet and corresponding to $u$.
    Total uncertainty is denoted by $2\bar{\sigma}$.
    }
    \label{fig:darcy_udeeponet}
\end{figure}

For Case B, we assume that a complete input $\lambda$ is available, i.e., a datapoint is available at each required location of $\lambda$.
We quantify model uncertainty in the prediction of $u$ using DEns. 
In particular, we run the DeepONet training algorithm five times  with different parameter initializations, and subsequently compute the predicted mean and uncertainties based on the five predictions of $u$; see Eqs.~\eqref{eq:uqt:pre:mcestmc:mean}-\eqref{eq:uqt:pre:mcestmc:totvar}. 
The error between the predicted mean and the reference solution of $u$ for an unseen $\log({\lambda})$ is bounded by the predicted uncertainties, as shown in Fig.~\ref{fig:darcy_udeeponet}.

\section{Summary}
\label{sec:conclusions}

Scientific machine learning (SciML) has emerged recently as an effective and powerful tool for data fusion, solving ordinary/partial differential equations (ODEs, PDEs), and learning operator mappings in various scientific and engineering disciplines. 
Indicative SciML models are the physics-informed neural networks (PINNs) for solving forward and inverse ODE/PDE problems and the deep operator networks (DeepONets) for learning operators that can, for instance, be used as fast solvers of ODEs/PDEs. 
In this context, quantifying predictive uncertainties is crucial for risk-sensitive applications as well as for efficient and economical design. 
In this regard, a comprehensive study of UQ methods for various SciML models, e.g., neural networks (NNs), PINNs, DeepONets, has been provided in \cite{psaros2022uncertainty}.

However, due to the fact that the implementations of current UQ methods are not straightforward, we have developed in this paper an open-source Python library 
(\href{https://github.com/Crunch-UQ4MI}{\textit{github.com/Crunch-UQ4MI}}),
termed NeuralUQ, that serves as an efficient and reliable toolbox for UQ in SciML.
In this study, we have presented a detailed tutorial of NeuralUQ, followed by four numerical examples, including dynamical systems and high-dimensional parametric and time-dependent PDEs, to demonstrate the applicability, reliability, and efficiency of NeuralUQ. We have further demonstrated the flexibility and customizability of NeuralUQ for supporting alternative surrogates, prior and posterior parameter distributions, as well as the inference methods. 
Indicative such modifications have been described and applied in regression and differential equation problems. More generally, in addition to the UQ methods presented herein for addressing the problem scenarios of Section~\ref{sec:uqINsciml:sciml}, NeuralUQ enables the users to address additional scenarios by combining the built-in components, such as the different surrogates and inference methods.

Despite the plurality of methods implemented in NeuralUQ, appropriately selecting a UQ method for solving a specific problem is not trivial. 
In this regard, we provide a few empirical remarks in the following. Deterministic methods, such as deep ensemble, are computationally efficient, and they provide reasonable model uncertainties for both neural ODE/PDE and neural operators problems. 
Further, they can handle big data by using data mini-batches, as in standard neural network (NN) training. 
For problems with noisy data, NN parameter regularization can be used for addressing overfitting. 
However, this may have a negative effect on predicted uncertainties.
Specifically, a large regularization weight can lead to small model uncertainty, whereas a small weight may be insufficient for addressing overfitting issues.  
As a potential remedy, different regularization weights can be used and compared by evaluating their corresponding predicted uncertainties using metrics and validation data. 
As for Bayesian methods, Hamiltonian Monte Carlo (HMC) typically achieves both satisfactory computational accuracy and efficiency for posterior estimation for problems with relatively small datasets, of the order of a few hundred measurements. For problems involving big data, the methods Langevin dynamics (LD), mean-field variational inference (MFVI), and Monte Carlo dropout (MCD), which support mini-batch training in a straightforward manner, are computationally more efficient. 
For solving neural ODE/PDE problems without and with historical data using Bayesian methods, we recommend the use of a BNN and the generator of a pre-trained physics-informed generative adversarial network as surrogates, respectively. 
Further, as a rule of thumb for neural operator problems, the physics-agnostic functional prior (PA-FP) can be used to quantify the uncertainties caused by incomplete data, and the deep ensemble for quantifying the epistemic uncertainties. 
Furthermore, the methods HMC and LD/MFVI are recommended for problems with small and big data, respectively, if FP is used.

Overall, we envision that NeuralUQ will be used in the future in large-scale computational science and engineering applications as a toolbox of reliable and efficient UQ implementations.
Clearly, such applications require up-to-date, well-maintained, and versatile software packages.
In this regard, indicative extensions of our library relate to incorporating more sampling methods that can handle big data, e.g., stochastic gradient Hamiltonian Monte Carlo, and variational inference methods with more accurate parameterized posterior distributions, e.g., normalizing flows. 
Finally, we will continue updating the library and we also encourage future users and researchers to develop additional components for NeuralUQ with the goal of expanding its current capabilities and advancing the field.


\section*{Acknowledgments}

This work is supported by the DOE PhILMs project (No. DE-SC0019453), the 
MURI-AFOSR FA9550-20-1-0358 projects, and the 
DARPA HR00112290029 project.

\bibliographystyle{siamplain}
\bibliography{main}

\section{Appendix}

In this section, we provide 
a glossary table in Appendix~\ref{sec:glossary}, and an example for surrogate customization, i.e., a one-dimensional regression problem, in Appendix ~\ref{sec:sine}. We then demonstrate the customization for prior in Samplable and posterior estimation in Variational, using a one-dimensional differential equation problem, in Appendix ~\ref{sec:1d_de}, and provide the details for data generation in Appendix~\ref{sec:training_details}.

\subsection{Glossary for frequently used terms}
\label{sec:glossary}

In Table~\ref{tab:glossary}, we provide 
a glossary table of frequently used terms in this paper, and in ML/SciML more generally.

\begin{table}[h]
    \centering
    \scriptsize
    \begin{tabular}{cccc}
    \toprule
	\multicolumn{4}{c}{\textbf{Glossary}}\\
	\midrule
    \bf{UQ} & uncertainty quantification & \bf{SciML} & scientific machine learning \\
    \midrule
    \bf{DNNs}& deep neural networks & \bf{BNNs} & Bayesian neural networks \\
    \midrule
    \bf{GANs}& generative adversarial networks & \bf{FNN} & fully-connected neural network  \\
    \midrule
   \bf{PINNs} & physics-informed neural networks & \bf{DeepONets} & deep operator networks\\
    \midrule
  \bf{FP} & functional prior & \bf{PI-GANs} & physics-informed GANs\\
      \midrule
   \bf{PA-FP} & physics-agnostic FP  & \bf{GAN-FP}  &  GAN-based FP \\
 \midrule
 \bf{BNN-FP} & BNN-based FP & & \\
    \bottomrule
    \end{tabular}
    \caption{
    Glossary of frequently used terms in this paper. 
    }
    \label{tab:glossary}
\end{table}

\subsection{Customization of surrogates}\label{sec:sine}

We demonstrate the surrogate customization, using a one-dimensional regression problem. The target function is: 
\begin{align}
    u = 1.5 \sin(11x), ~ x \in [-1, 1].
\end{align}
We assume that we have 3 noisy measurements of $u$, which are equidistantly distributed in $[-0.7, -0.3]$. 
The noise for the measurements is assumed to be Gaussian with zero mean and known standard deviation set to 0.05.  

We employ three different surrogates for different scenarios: (1) We have neither historical data nor prior knowledge on the formulation of the target function. 
We then employ a BNN with one hidden layer which has 50 neurons and the hyperbolic tangent activation function as the surrogate. 
In addition, the standard normal distribution is utilized as the prior distribution for each weight and bias in the BNN; 
(2) the expression describing the target function is not considered as known, but we have historical data for it. 
We then train a GAN to learn the functional prior using historical data; and (3) we have prior knowledge regarding the target function expressing that it is a sine function with unknown amplitude and frequency, i.e., $A\sin(\omega x)$, where $A$ and $\omega$ are parameters to be inferred given measurements.
The details regarding the pre-trained GAN in (2) can be seen in the GitHub repository, and the posteriors for the parameters in each surrogate are computed using HMC.

\begin{figure}[h]
    \centering
    \subfigure[]{
    \includegraphics[width=0.3\textwidth]{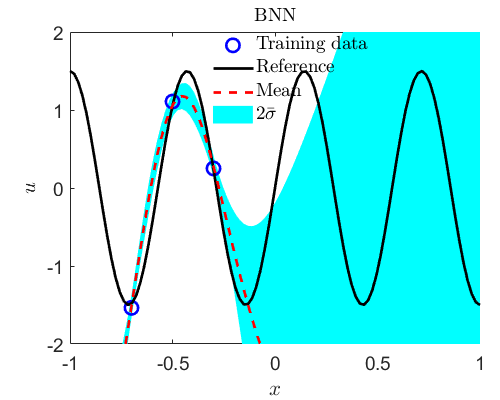}
    }
    \subfigure[]{
    \includegraphics[width=0.3\textwidth]{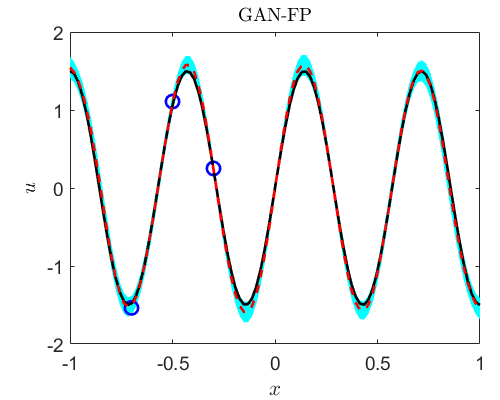}
     }
    \subfigure[]{
    \includegraphics[width=0.3\textwidth]{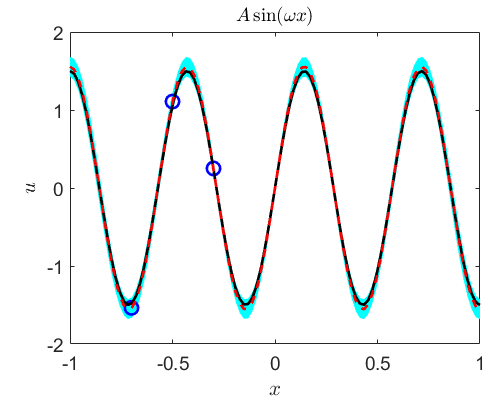}
    }
    \caption{
    1D regression problem: Predictions for $u$ using different surrogates. (a) BNN; (b) GAN-FP, and (c) $A\sin(\omega x)$.  $2\bar{\sigma}$: Total uncertainty.
    }
    \label{fig:sine}
\end{figure}


As shown in Fig.~\ref{fig:sine}, the predictions from the GAN-FP and $A\sin({\omega}x)$ surrogates are quite similar for both the predicted means and uncertainties, which are more accurate than the results from the BNN surrogate in terms of the means. 
Also, the errors between the predicted means and the reference solution are mostly bounded by the predicted uncertainties from GAN-FP and $A\sin({\omega}x)$, whereas the errors are only partly bounded by the predicted uncertainties in BNN. 
Overall, we demonstrate the flexibility and customizability of NeuralUQ by customizing problem-dependent surrogates to achieve better predicted accuracy. 

\subsection{Customization of prior and inference}\label{sec:1d_de}

We demonstrate the customization for prior in Samplable
and posterior estimation in Variational, using a one-dimensional differential equation problem. Specifically, we consider a nonlinear diffusion-reaction system expressed as follows:
\begin{align}\label{eq:1d_de}
    D\partial_x^2 u - k_r u^3 = f,~ x\in[-1, 1],
\end{align}
where $u$ represents the concentration of a certain solute, $D$ denotes the diffusion coefficient, $k_r$ is the reaction rate, and $f$ is the source term. 

We first demonstrate the customization of prior using the example of an inverse problem described by Eq. \eqref{eq:1d_de}. 
In this problem, $D = 0.01$ is a known constant, and we have a few noisy measurements on $u$, and $f$. The objective is to infer $u$ and $f$ for $x \in [-1, 1]$, and the reaction rate $k_r$ which is an unknown constant. 
Specifically, we have noisy measurements of $u$ at 5 random locations, and 17 noisy equidistant measurements of $f$. We then employ a BNN as well as a {Samplable} variable as the surrogate for $u$ and $k_r$, respectively. The equation can then be encoded using the automatic differentiation as in PINNs. The employed BNN has 3 hidden layers with 50 neurons per layers, and the  hyperbolic tangent activation function is used as the activation function. We assume that each weight/bias has the same prior distribution in this BNN, which is independent Gaussian here, i.e.,  $N(0, 1^2)$. Further, we employ three different prior distributions for $k_r$, i.e., $k_r \sim N(0, 1^2)$, $k_r \sim HalfN(0.5, 0.5^2)$ and $\log k_r \sim N(0, 1^2)$, where $HalfN$ refers to half-normal distribution. 
The posterior samples for the parameters in BNN as well as $k_r$ are obtained using HMC. We only present the predictions for $k_r$ in Table~\ref{tab:1d_de_inverse}, for simplicity. As shown, the predictions for $k_r$ are strongly affected by its prior distribution. Overall, NeuralUQ enables the users to utilize different prior distributions to achieve better accuracy.

\begin{table}[h]
    \footnotesize
    \centering
    \begin{tabular}{c|ccc}
    \toprule
        Priors for $k_r$ & $ k_r \sim N(0, 1^2)$ & $k_r \sim HalfN(0, 1^2)$ & $\log k_r \sim N(0, 1^2)$ \\
         \midrule
       mean $\pm$ std & 0.5727 $\pm$ 0.4407 & 0.4131 $\pm$ 0.2488 & 0.4538 $\pm$ 0.2550 \\
       \bottomrule
    \end{tabular}
    \caption{
    Nonlinear diffusion-reaction problem: Predictions for $k_r$ with different priors. The reference solution is $k_r = 0.2$.
    }
    \label{tab:1d_de_inverse}
\end{table}

\begin{figure}[h]
    \centering
    \subfigure[]{
    \includegraphics[width=0.35\textwidth]{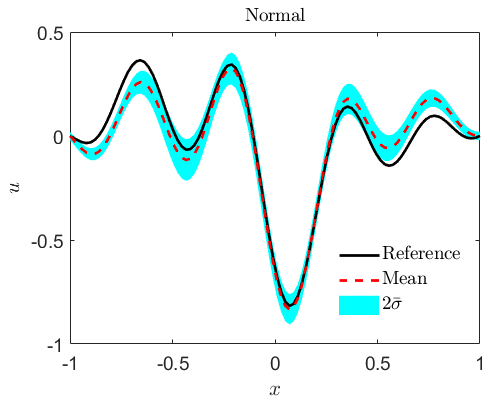}
    \includegraphics[width=0.35\textwidth]{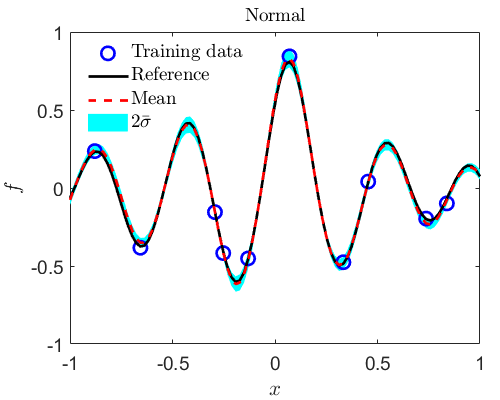}
    }
    \subfigure[]{
    \includegraphics[width=0.35\textwidth]{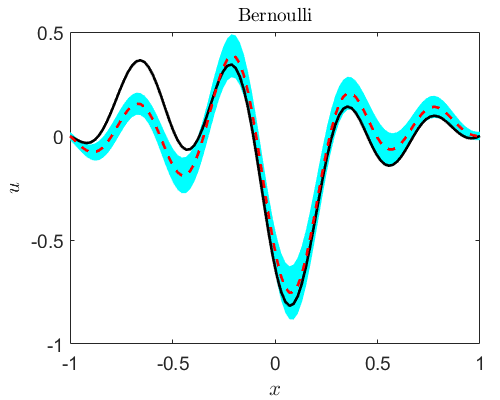}
    \includegraphics[width=0.35\textwidth]{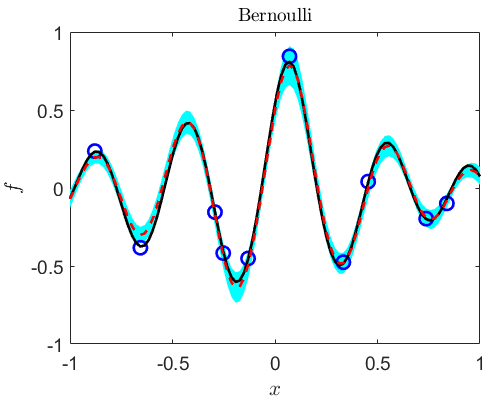}
    }
    \caption{
    Nonlinear diffusion-reaction problem: Predictions for $u$ and $f$ from VI with different parameterized posteriors. (a) Mean-field Gaussian, (b) Mean-field Bernoulli.
    }
    \label{fig:FP_1D}
\end{figure}

We proceed to show the customization of posterior estimate in Variational using the example of a forward problem governed by Eq. \eqref{eq:1d_de}. 
In particular, $D = 0.01$, $k_r = 0.2$, and the boundary condition is $u(-1) = u(1) = 0$. In addition, we assume that we have 10 noisy measurements of $f$ which are randomly distributed at $x \in [-1, 1]$. The noise for the measurements, which is $N(0, 0.01^2)$, is assumed to be known. The objective is to infer $u$ given data of $f$. We employ a PI-GAN, which is the same as in \cite{meng2022learning}, to solve this problem. The same historical data in \cite{meng2022learning} is utilized here to train the PI-GAN to obtain the functional priors for $u$ and $f$. Note that HMC is employed in \cite{meng2022learning} for the posterior estimation of the parameters in PI-GAN. Here we utilize VI with two different parameterized posterior distributions to demonstrate the flexibility of NeuralUQ, i.e., independent Gaussian with unknown means and standard deviations (mean-field Gaussian), and independent Bernoulli distributions with fixed $p=0.5$ as the probability of success/failure, multiplied by unknown variables (mean-field Bernoulli). The predictions for $u$ and $f$ are displayed in Fig.~\ref{fig:FP_1D}, it is observed that (1) the predicted means for $u$ and $f$ from the two parameterized posteriors are similar, and (2) the errors between the predicted means for $u$ and $f$ are mostly bounded by the predicted uncertainties in both methods.


Finally, note that we employ these specific prior/posterior distributions for demonstration purposes. Interested users of NeuralUQ can also customize other problem-dependent prior/posterior distributions by following the examples presented above.

\subsection{Data generation}
\label{sec:training_details}

Regarding data generation, in Section~\ref{example:ko}, the nonstiff differential equations solver \textit{ode45} in Matlab is utilized for solving Eq.~\eqref{eq:ko} to generate the training data as well as the reference solutions; in Section~\ref{example:kdv}, the training data are generated from the exact solution of $u$ from \cite{benes2006decompositions}; and in Section~\ref{example:darcy}, the finite-element-based Partial Differential Equation Toolbox in Matlab is utilized for solving Eq.~\eqref{eq:darcy_eq} to generate the training data as well as the reference solutions. Note that all data as well as the employed NNs related to this study can be accessed in the GitHub repository of NeuralUQ.

\end{document}